\documentclass{article}
\usepackage{ijcai22}

\usepackage{times}
\usepackage{soul}
\usepackage{url}
\usepackage[hidelinks]{hyperref}
\usepackage[utf8]{inputenc}
\usepackage[small]{caption}
\usepackage{graphicx}
\usepackage{amsmath}
\usepackage{amsthm}
\usepackage{booktabs}
\usepackage{algorithm}
\usepackage{algorithmic}
\usepackage{color}
\usepackage{subfigure}
\usepackage{amssymb}
\newtheorem{theorem}{Theorem}
\newtheorem{definition}{Definition}
\newtheorem{assumption}{Assumption}

\newtheorem{remark}{Remark}

\newcommand{\E}{{\mathbb{E}}}

\title{Bridging Differential Privacy and Byzantine-Robustness via Model Aggregation}

\author{
Heng Zhu$^{1,2}$
\and
Qing Ling$^1$
\affiliations
$^1$Sun Yat-sen University\\
$^2$University of California, San Diego
\emails
hez007@ucsd.edu,
lingqing556@mail.sysu.edu.cn
}

\begin{document}

\maketitle

\begin{abstract}
This paper aims at jointly addressing two seemly conflicting issues in federated learning: differential privacy (DP) and Byzantine-robustness, which are particularly challenging when the distributed data are non-i.i.d. (independent and identically distributed). The standard DP mechanisms add noise to the transmitted messages, and entangle with robust stochastic gradient aggregation to defend against Byzantine attacks. In this paper, we decouple the two issues via robust stochastic model aggregation, in the sense that  our proposed DP mechanisms and the defense against Byzantine attacks have separated influence on the learning performance. Leveraging robust stochastic model aggregation, at each iteration, each worker calculates the difference between the local model and the global one, followed by sending the element-wise signs to the master node, which enables robustness to Byzantine attacks. Further, we design two DP mechanisms to perturb the uploaded signs for the purpose of privacy preservation, and prove that they are $(\epsilon,0)$-DP by exploiting the properties of noise distributions. With the tools of Moreau envelop and proximal point projection, we establish the convergence of the proposed algorithm when the cost function is nonconvex. We analyze the trade-off between privacy preservation and learning performance, and show that the influence of our proposed DP mechanisms is decoupled with that of robust stochastic model aggregation. Numerical experiments demonstrate the effectiveness of the proposed algorithm.
\end{abstract}

\section{Introduction}

Federated learning has attracted much attention from both industry and academia at the time of rapid development of distributed intelligent devices. At each iteration of a federated learning algorithm, a master node aggregates the local information sent from workers (namely, distributed intelligent devices) to update a global model. The local data of each worker are kept private, and do not need to be shared with other workers or the master node \cite{konevcny2016federated,MAL-083}. In the process of transmitting the local information, such as stochastic gradients or model parameters, there are several challenges to be addressed, including data privacy, robustness to malicious attacks, and communication efficiency. In this paper, we are particularly interested in the privacy and robustness issues. In a nutshell, privacy and robustness are to handle two different threats. Privacy concerns the threat from the \textit{curious but honest} master node, who potentially expects to recover the private local data from the transmissions of local information. Robustness concerns the threat from the \textit{dishonest and adversarial} workers, who aim at biasing the learning process.

%\cite{konevcny2016federated,yang2019federated,MAL-083,wang2021field}

In privacy-preserving data analysis, differential privacy (DP) \cite{dwork2014algorithmic} is a gold standard and has a wide range of applications. In the popular parameter server architecture and distributed stochastic gradient descent (SGD) algorithm, adding Gaussian noise to the transmitted stochastic gradients is a common approach to achieve DP \cite{song2013stochastic,abadi2016deep,wei2020federated}. The level of privacy guarantee is tuned by the variance of added noise. For adversarial behaviors, we consider the Byzantine attacks model to characterize that some workers may send faulty messages to bias the aggregation at the master node \cite{LamportSP82,chen2017distributed,guerraoui2018hidden,MAL-083}. Byzantine attacks are devastating to the distributed SGD algorithm, in which the master node uses mean aggregation for the local stochastic gradients. A common remedy is to replace mean aggregation with other robust aggregation rules, such as geometric median, median, trimmed mean, etc \cite{chen2017distributed,blanchard2017machine,yin2018byzantine,wu2020federated,karimireddy2020learning}. However, the noise of stochastic gradients from regular workers will weaken the ability of defending against Byzantine attacks since larger variance of stochastic gradients will make the elimination of malicious messages much harder \cite{wu2020federated,karimireddy2020learning}. Thus, the added noise of DP mechanisms shall harm the performance of robust stochastic gradient aggregation, resulting the conflict between privacy preservation and defense against Byzantine attacks. For example, the added Gaussian noise can make the regular stochastic gradients undistinguishable with the malicious messages from the Byzantine workers. The work of \cite{guerraoui2021differential} formally analyzes the incompatibility of existing robust stochastic gradient aggregation rules and DP mechanisms, and \cite{guerraoui2021combining} further shows the multiplicative influence of robust stochastic gradient aggregation and DP mechanisms on the learning performance.

%\cite{song2013stochastic,abadi2016deep,arachchige2019local,wei2020federated}

In this paper, we tackle the problem by decoupling the two issues via model aggregation rather than the common gradient aggregation. We propose a Differentially-Private Robust Stochastic model Aggregation (DP-RSA) algorithm to jointly address the privacy and robustness issues in federated learning, where the workers have non-i.i.d. data. At each iteration, each worker calculates the difference between the local model and the global one, followed by sending the element-wise signs to the master node, which enables defense against Byzantine attacks over non-i.i.d. data. Adaptive to robust stochastic model aggregation, we design two DP mechanisms, Sign-Flipping and Sign-Gaussian, to perturb the uploaded signs for the purpose of privacy preservation. By theoretical analysis, we display the trade-off between privacy preservation and learning performance, and
point out that the separated, additive influence of our proposed DP mechanisms and robust stochastic model aggregation. Beyond provable privacy preservation and Byzantine-robustness, DP-RSA also enjoys favorable communication efficiency as the workers only send signs to the master node.

Proving privacy preservation and Byzantine-robustness for DP-RSA is challenging. To show the proposed DP mechanisms satisfy $(\epsilon, 0)$-DP, we have to investigate the impact of added noise to the signs. In particular, for the Sign-Gaussian mechanism, analyzing the impact of added Gaussian noise on the signs is difficult, and we address it by exploiting the cumulative distribution function (CDF) of Gaussian distribution. To show the convergence of DP-RSA, we must handle the nonconvex and nonsmooth cost function, for which common measures of convergence are not applicable. We leverage Moreau envelop and proximal point projection \cite{davis2019stochastic} to establish the convergence under the assumption of weak convexity.

Our contributions are summarized as follows.
\begin{itemize}
    \item We propose a Differentially-Private Robust Stochastic model Aggregation (DP-RSA) algorithm for federated learning over distributed non-i.i.d. data, simultaneously meeting the requirements of privacy preservation, Byzantine-robustness, and communication efficiency.
    \item We design two DP mechanisms, Sign-Flipping and Sign-Gaussian, to perturb the uploaded signs for the purpose of privacy preservation. We rigorously prove that both mechanisms satisfy $(\epsilon, 0)$-DP. The proofs can be extended to other DP mechanisms that involve signs.
    \item We prove the convergence of DP-RSA and point out the additive impact of DP mechanisms and robust stochastic model aggregation on the learning performance. For the nonconvex and nonsmooth cost function, we leverage Moreau envelop and proximal point projection to establish the convergence. The convergence analysis is novel in the context of robust nonconvex distributed learning.
\end{itemize}

\section{Problem Formulation}

Consider a distributed federated learning system with one master node and $K$ workers. Among these workers, $r$ of them are regular and constitute a set $\mathcal{R}$, while the rest $b$ of them are Byzantine and constitute a set $\mathcal{B}$, where $K = r+b$. Note that the numbers and identities of regular and Byzantine workers are unknown to the master node. The Byzantine workers are assumed to be omniscient and can collude with each other to send arbitrary malicious messages to the master node. The problem of interest is to find an acceptable solution to the nonconvex distributed learning problem
\begin{equation}
\label{problem1}
    \min\limits_{\tilde{x} \in \mathbb{R}^d} \sum\limits_{k \in \mathcal{R}} E[f_k(\tilde{x}, \zeta_k)] + f_0(\tilde{x}),
\end{equation}
where $\tilde{x} \in \mathbb{R}^d$ is the model to be optimized, $f_k(\tilde{x}, \zeta_k)$ is the nonconvex local cost function at worker $k$ with respect to a random variable $\zeta_k$, and $f_0(\tilde{x})$ is a regularization term at the master node. In this work we consider a non-i.i.d. setting for the distributed data. That is to say, the random variables $\zeta_k \sim D_k$, where $D_k$ represent the data distributions at workers $k$ and they can be different to each other at different workers. Note that non-i.i.d. data distribution is common in federated learning, and brings remarkable challenges for designing Byzantine-robust algorithms.

\subsection{Differential Privacy (DP)}

The definition of differential privacy (DP) is as follows.

\begin{definition}
A randomized algorithm $\mathcal{M}$ is $(\epsilon, \delta)$-DP if for all $\mathcal{I}_a, \mathcal{I}_b$ that are a pair of adjacent inputs $\left \| \mathcal{I}_a - \mathcal{I}_b \right \|_1 \leq 1$ and for all possible set of outputs $\mathcal{O}$, it holds
\begin{equation}
    \mathrm{Pr}[\mathcal{M}(\mathcal{I}_a) \in \mathcal{O}] \leq e^\epsilon \mathrm{Pr}[\mathcal{M}(\mathcal{I}_b) \in \mathcal{O}] + \delta.
\end{equation}
\end{definition}
Here $(\epsilon, \delta)$ represents the privacy budget to guarantee DP. The privacy loss $\epsilon$ controls the trade-off between privacy and utility of the algorithm, and $\delta$ is the probability that $\epsilon$-DP can fail. In a federated learning system, at each iteration the workers send their local messages to the master node once. Thus, a local randomizer should be used to perturb the local message sent from each worker. In the context of distributed learning, we analyze the per-round-of-communication privacy budget $(\epsilon, \delta)$ to ensure privacy, which is also the case in the works of, for example, \cite{agarwal2018cpsgd,jin2020stochastic,guerraoui2021differential}. Indeed, it is also doable to exploit advanced composition theorems \cite{dwork2014algorithmic,kairouz2015composition} or the analytical moments accountant method \cite{abadi2016deep,mironov2017renyi} to obtain the multi-round privacy loss.

\section{Algorithm Development}
In this section we develop an algorithm that jointly addresses the privacy and robustness issues.
%As mentioned above, the performance of robust stochastic gradient aggregation becomes worse when considering DP, since the added noise makes the defense against Byzantine attacks much harder.
We adopt the idea of robust stochastic model aggregation to handle the attacks on the non-i.i.d. data distribution, and develop DP mechanisms to protect data privacy.

\subsection{Byzantine-Robustness via Model Aggregation}
We first introduce robust stochastic model aggregation to defend against Byzantine attacks \cite{li2019rsa}. Note that (\ref{problem1}) can be rewritten as
\begin{align}
\label{problem2}
     \min\limits_{x} & \sum\limits_{k \in \mathcal{R}} E [f_k(x_k, \zeta_k)] + f_0(x_0), \\
     \text{s.t.} ~ & ~ x_0 = x_k, ~ \forall k \in \mathcal{R}, \notag
\end{align}
where $x := [\cdots; x_k; \cdots; x_0] \in \mathbb{R}^{(r+1)d}$ consists of $r$ local models $x_k$ at all regular workers and $x_0$ at the master node. Intuitively, the local models should be the same no matter the distributed data are non-i.i.d. or not -- this is different to the local stochastic gradients. To enable robust stochastic model aggregation, we use an $\ell_1$-norm penalty term to relax the constraints in (\ref{problem2}), as
\begin{equation}
\label{problem3}
    \min\limits_{x} \sum\limits_{k \in \mathcal{R}} \left( E [f_k(x_k, \zeta_k)] + \lambda \left \| x_k - x_0 \right \|_1 \right) + f_0(x_0),
\end{equation}
where the $\ell_1$-norm penalty parameterized by $\lambda > 0$ forces the local variables $x_k$ to be close to $x_0$, but allows them to be different for the sake of tolerating Byzantine attacks.

When there are no Byzantine workers, we can apply the stochastic subgradient method with step size $\alpha^t > 0$ to solve (\ref{problem3}). With $g_k^t := \nabla f_{k}(x_k^t, \zeta_k^t)$, at iteration $t+1$ we have
\begin{align}
    \label{updatek}
    x_k^{t+1} & = x_k^t - \alpha^t \left( g_k^t + \lambda \text{sign} (x_k^t - x_0^t) \right), \\
    \label{update0}
    x_0^{t+1} & = x_0^t - \alpha^t \left( \nabla f_0(x_0) + \lambda \sum_{k \in \mathcal{R}} \text{sign} (x_0^t - x_k^t) \right),
\end{align}
where the element-wise function $\text{sign}(\cdot)$ returns $1$ for nonnegative input and $-1$ for negative input. At iteration $t+1$, the master node first broadcasts the model $x_0^t$ to all workers. The regular workers $k$ update their local models as (\ref{updatek}), and send back $\text{sign} (x_0^t - x_k^t)$ to the master node. Upon receiving all local messages, the master node updates $x_0^{t+1}$ as (\ref{update0}).

In the presence of Byzantine workers $j \in \mathcal{B}$, they can generate arbitrary vectors $z_j^t \in \mathbb{R}^d$ and send $\text{sign}(x_0^t - z_j^t)$  to the master node. Thus, the regular workers still update the local models as (\ref{updatek}), but the master node updates $x_0^{t+1}$ as
\begin{align}\label{update0-new}
    x_0^{t+1}  = x_0^t - \alpha^t & \Big( \nabla f_0(x_0) + \lambda \Big( \sum\limits_{k \in \mathcal{R}} \text{sign} (x_0^t - x_k^t) \notag \\
    &  + \sum\limits_{j \in \mathcal{B}} \text{sign} (x_0^t - z_j^t) \Big) \Big).
\end{align}

As shown in \cite{li2019rsa}, with robust stochastic model aggregation, each regular or Byzantine worker has the same impact on the model update at the master node (that is, $\alpha^k \lambda$ per element), regardless of the actual vector generated. As a consequence, the negative effects brought by the malicious messages are only relative to the number of Byzantine workers, not to the values of the malicious messages.

%That is why the $\ell_1$-norm penalty helps establish the Byzantine-robust method.

\subsection{DP Mechanisms}

However, the robust stochastic model aggregation updates \eqref{updatek} and \eqref{update0-new} have the risk of leaking private data, since each regular worker $k$ still needs to send $\text{sign}(x_0^t - x_k^t)$ to the master node. To guarantee DP, we must introduce randomness in the transmitted signs.
%We use differential privacy mechanisms to guarantee the privacy of the transmitted information from workers.
Here we propose two DP mechanisms adaptive to robust stochastic model aggregation.

%In the robust model aggregation, at time $t$, one regular worker sends $\text{sign}(x_0^t - x_k^t)$ to the master node.

\subsubsection{Sign-Flipping Mechanism}
We first propose a straightforward mechanism to introduce randomness in the transmitted signs. The $i$-th element of vector $\text{sign}(x_0^t - x_k^t)$ flips its sign with probability $1-\gamma$, as
\begin{align}
\label{flip}
    & \text{Flip}(\text{sign}(x_0^t - x_k^t)_i) = \notag \\
    & \left \{
    \begin{array}{rrl}
         & -\text{sign}(x_0^t - x_k^t)_i, & ~ \text{with probability} ~ 1 - \gamma, \\
         & \text{sign}(x_0^t - x_k^t)_i,  & ~ \text{with probability} ~ \gamma.
    \end{array} \right.
\end{align}
After receiving the signs from the workers, the master node cannot exactly identify the actual signs, which protects data privacy to some extent.

\subsubsection{Sign-Gaussian Mechanism}
Motivated by the conventional Gaussian mechanism in DP, we add Gaussian noise to the model difference $x_0^t - x_k^t$ and then obtain the signs. At iteration $t+1$, each regular worker $k$ sends $\text{sign}(x_0^t - x_k^t + e_k^t)$ to the master node, where $e_k^t \sim \mathcal{N}(0, \sigma^2 I_d) \in \mathbb{R}^d$ is the multivariate Gaussian noise with zero mean and $\sigma^2$ variance, and $I_d$ is the $d \times d$ identity matrix. Thus, the signs can be randomly changed after adding the noise. We write the Sign-Gaussian mechanism as
\begin{align}
\label{gaussian}
    \text{Gaussian}(\text{sign}(x_0^t - x_k^t)) = \text{sign} (x_0^t - x_k^t + e_k^t).
\end{align}
%Let $u_k^t = x_0^t - x_k^t$, $\Delta u$ be the $\ell_2$-norm sensitivity of $u_k^t$, and $y_k^t = \text{sign} (x_0^t - x_k^t + e_k^t)$ be the output of the Sign-Gaussian mechanism, where $y_k^t \in \{1,-1\}$. We can obtain its probability distribution as
%\begin{equation}
%    \text{Pr}(y_k^t | u_k^t) = \Phi (\frac{y_k^t u_k^t}{\sigma})
%\end{equation}
%where $\Phi (\cdot)$ is the cumulative distribution function (cdf) of the standard Gaussian distribution, i.e. $\Phi (x) = \frac{1}{\sqrt{2\pi}} \int_{-\infty}^x e^{- t^2/2} dt$.
Let $u_k^t = x_0^t - x_k^t$ and $y_k^t = \text{sign} (x_0^t - x_k^t + e_k^t)$ be the output of the Sign-Gaussian mechanism, where each element of $y_k^t$ belongs to  $\{1,-1\}$. We can obtain the probability distribution of the $i$-th element of $y_k^t$ as
\begin{equation}
    \text{Pr} ((y_k^t)_i | (u_k^t)_i) = \Phi \left( \frac{(y_k^t)_i (u_k^t)_i}{\sigma} \right),
\end{equation}
where $\Phi (\cdot)$ is the CDF of the standard normal distribution, given by $\Phi (a) = \frac{1}{\sqrt{2\pi}} \int_{-\infty}^a e^{- s^2/2} ds$.

\subsection{DP-RSA Algorithm}
Based on robust stochastic model aggregation and the two DP mechanisms, we propose the DP-RSA algorithm, which achieves privacy-preserving and Byzantine-robust federated learning over non-i.i.d. data. For notational convenience, we use $F/G(\cdot)$ to denote either the Flip function in (\ref{flip}) or the Gaussian function in (\ref{gaussian}).

The algorithm is described as Algorithm \ref{algorithmdp}. At iteration $t+1$, the master node broadcasts the model $x_0^t$ to all workers. The regular workers $k \in \mathcal{R}$ update their local models, and send back the perturbed signs $F/G(\text{sign}(x_0^t - x_k^t))$ to the master node. The Byzantine workers $j \in \mathcal{B}$ generate arbitrary vectors $z_j^t$ and send $F/G(\text{sign}(x_0^t  - z_j^t))$ to the master node. With particular note, for the Byzantine workers $j \in \mathcal{B}$, $F/G(\text{sign}(x_0^t  - z_j^t))$ and $\text{sign}(x_0^t  - z_j^t)$ are essentially equivalent. Upon receiving the signs from all workers, the master node updates $x_0^{t+1}$ as
\begin{align}
\label{updatedp0}
    x_0^{t+1}  = x_0^t - \alpha^t & \Big( \nabla f_0(x_0) + \lambda \Big( \sum\limits_{k \in \mathcal{R}} F/G (\text{sign} (x_0^t - x_k^t)) \notag \\
    & + \sum\limits_{j \in \mathcal{B}} \text{sign} (x_0^t - z_j^t) \Big) \Big).
\end{align}

The robust stochastic model aggregation rule is able to defend against Byzantine attacks even for non-i.i.d. data. The DP mechanisms can ensure data privacy during the training process. In addition, the transmissions of signs are communication-efficient. Thus, our proposed DP-RSA simultaneously meets the requirements of privacy preservation, Byzantine-robustness, and communication efficiency.

\begin{remark}
It is of interest to observe that the Sign-Flipping mechanism shares similarity with Sign-Flipping Attacks, while the Sign-Gaussian mechanism is close to Gaussian Attacks. These observations indicate the trade-off between privacy preservation and learning performance.
%The privacy loss should be controlled at a reasonable level to achieve an acceptable learning error.
In addition, since the uploaded signs of all workers have equal contributions to the model update at the master node, the impact of attacks from Byzantine workers is not coupled with that of DP mechanisms used in regular workers. We will characterize these observations in the ensuing theoretical analysis.
\end{remark}

\begin{algorithm}[tb]
\caption{DP-RSA}
\label{algorithmdp}
\textbf{Input}: Step size $\alpha^t$, penalty parameter $\lambda$, hyperparameter $\gamma$ in Sign-Flipping or $\sigma$ in Sign-Gaussian \\
\textbf{Initialize}: Initialize $x_0^0$ for master node \\
\vspace{-1.2em}
\begin{algorithmic}[1] %[1] enables line numbers
\FOR {$t=0,1,\dots$}
\STATE \textbf{Master Node}:
\STATE Broadcast $x_0^t$ to all workers
\STATE Receive $F/G(\text{sign}(x_0^t - x_k^t))$ from regular workers and $F/G(\text{sign}(x_0^t - z_j^t))$ from Byzantine workers
\STATE Update $x_0^{t+1}$ as (\ref{updatedp0})
\STATE \textbf{Worker $k$ or $j$}:
\IF {$k \in \mathcal{R}$}
\STATE Receive $x_0^t$ from master node
\STATE Send $F/G(\text{sign}(x_0^t - x_k^t))$ to master node
\STATE Randomly select samples $\zeta_k^t \sim D_k$ and obtain stochastic gradient $g_k^t = \nabla f_k(x_k^t, \zeta_k^t)$
\STATE Update local model $x_k^{t+1}$
\ELSIF {$j \in \mathcal{B}$}
\STATE Receive $x_0^t$ from master node
\STATE Generate arbitrary malicious vector $z_j^t$
\STATE Send $F/G(\text{sign}(x_0^t - z_j^t))$ to master node
\ENDIF
\ENDFOR
\end{algorithmic}
%\textbf{Output}: $x$
\end{algorithm}

\section{Theoretical Analysis}
In this section, we theoretically analyze the DP-RSA algorithm. We first prove the proposed Sign-Flipping and Sign-Gaussian mechanisms satisfy $(\epsilon, 0)$-DP. Then we analyze the convergence of DP-RSA for the non-convex problem.

%\subsection{Assumptions}
Now we give several assumptions used in the analysis. For notational convenience, define $f_k(\tilde{x}):=E[f_k(\tilde{x}, \zeta_k)]$ as the local cost function of regular worker $k$.

\begin{assumption}[Weakly Convexity]
\label{a2}
The local cost functions $f_k(\tilde{x})$ of regular workers $k$ and the regularization term $f_0(\tilde{x})$ are $\rho$-weakly convex, which implies $f_k(\tilde{x}) + \frac{\rho}{2} \left \| \tilde{x} \right \|^2$ and $f_0(\tilde{x}) + \frac{\rho}{2} \left \| \tilde{x} \right \|^2$ are convex. For any $\tilde{x},\tilde{y} \in \mathbb{R}^d$, it holds
\begin{align}
    f_k(\tilde{y}) & \geq f_k(\tilde{x}) + \left< \nabla f_k(\tilde{x}), \tilde{y}-\tilde{x} \right> - \frac{\rho}{2} \left \| \tilde{y}-\tilde{x} \right \|^2,  \\
    & \quad ~ \forall k \in \mathcal{R} \cup 0. \notag
\end{align}
\end{assumption}

\begin{assumption}[Bounded Gradient]
\label{a3}
For any regular worker $k \in \mathcal{R}$ and any $x_k^t \in \mathbb{R}^d$, the stochastic gradient $g_k^t := \nabla f_{k}(x_k^t, \zeta_k^t)$ is upper-bounded by
\begin{equation}
    E \left \| g_k^t \right \|^2 \leq M^2.
\end{equation}
For any $x_0^t \in \mathbb{R}^d$, the gradient $\nabla f_0(x_0^t)$ at the master node is also upper-bounded by
\begin{equation}
    \left \| \nabla f_0(x_0^t) \right \|^2 \leq M^2.
\end{equation}
\end{assumption}

Assumption \ref{a2} relaxes the assumption of convexity. In the fields of statistical learning and signal processing, weakly convex functions are common, such as nonlinear least squares, phase retrieval, robust principal component analysis, and so on. For example, a function in the form of $f_1(\cdot) + f_2(\cdot)$, where $f_1(\cdot)$ has $\rho$-Lipschitz continuous gradient and $f_2(\cdot)$ is closed and convex, is $\rho$-weakly convex \cite{davis2019proximally}. Assumption \ref{a3} is common in differentially private machine learning and introduced to control the sensitivity. This assumption is natural in deep learning since gradient clipping is a standard operation to constrain the gradient norms.

\subsection{DP Guarantee}

For the Sign-Flipping mechanism, it is straightforward to show that it satisfies $(\epsilon, 0)$-DP; see the extended version of this paper \cite{zhu2022bridging}.
\begin{theorem}
\label{theoremflip}
The Sign-Flipping operation given by (\ref{flip}) satisfies $(\ln \frac{\gamma}{1-\gamma}, 0)$-DP.
\end{theorem}

Next, we prove that the proposed Sign-Gaussian mechanism satisfies $(\epsilon, 0)$-DP.
%\blue{Although the DP-Sign mechanism in \cite{jin2020stochastic} is similar to the Sign-Gaussian mechanism, our proof exploits the CDF of Gaussian distribution and is new.}
%Here we give a new proof that sign-Gaussian mechanism actually satisfies $(\epsilon, 0)$-DP.
We use $u_k^t$ and $u_k^{t \prime}$ to represent the outputs of two adjacent datasets and $v_k^t = u_k^{t \prime} - u_k^t$. The vector $v_k^t$ satisfies $ \left \|v_k^t \right \| \leq \Delta u$, where $\Delta u$ is the $\ell_2$-norm sensitivity of $u_k^t$. To measure the privacy loss (PL) in DP, we consider
\begin{align}
    PL & = \ln \frac{\text{Pr}(y_k^t | u_k^t)}{\text{Pr}(y_k^t | u_k^t + v_k^t)} \\
    = & \sum\limits_{i=1}^d \ln \frac{\Phi( (y_k^t)_i (u_k^t)_i / \sigma)}{\Phi( (y_k^t)_i ((u_k^t)_i + (v_k^t)_i)/\sigma)} \notag \\
     = & \sum\limits_{i=1}^d \left[ \ln \Phi(\frac{(y_k^t)_i (u_k^t)_i}{\sigma} ) - \ln \Phi( \frac{(y_k^t)_i ((u_k^t)_i + (v_k^t)_i)}{\sigma}) \right]. \notag
\end{align}
Observe that the function $\ln \Phi(\cdot)$ is crucial in privacy analysis. Actually its derivative is related to Mill's ratio \cite{sampford1953some,pinelis2019exact}, which can be used to characterize $\ln \Phi(\cdot)$. Based on the property of Mill's ratio, we give a novel proof of the Sign-Gaussian mechanism. The proof is left to \cite{zhu2022bridging}.

\begin{theorem}
\label{theoremgauss}
If the variance $\sigma^2$ of added Gaussian noise satisfies $\sigma > \max \{ \max_i \frac{2(u_k^t)_i}{3}, \frac{4\Delta u}{{\epsilon}} \}$, then the Sign-Gaussian operation given by \eqref{gaussian} satisfies $(\epsilon, 0)$-DP, where $\epsilon \in (0,8)$ is a constant.
\end{theorem}

As mentioned in Definition 1, the constant $\delta$ in $(\epsilon, \delta)$-DP represents the probability that $\epsilon$-DP fails, which is zero in our analysis.
%When two adjacent datasets are with at most one different sample,
When we use the constant step size $\alpha^t = \alpha$, $\Delta u$ can be set as $2 \alpha  M$. Observe that as the algorithm evolves, the distance between the local and global models can be closer, resulting smaller variance of added Gaussian noise.

Note that \cite{jin2020stochastic} proposes a similar DP-Sign operation in the SignSGD algorithm, and proves that it satisfies $(\epsilon, \delta)$-DP when the noise variance $\sigma^2$ is properly chosen. However, the proof in \cite{jin2020stochastic} just follows that for the conventional Gaussian mechanism, and hence ensures the same level privacy guarantee. In contrast, our proof exploits the CDF of Gaussian distributions, and leads to $\delta = 0$. Our proof techniques can also be applied in analyzing other DP algorithms involving signs.

\subsection{Convergence Analysis}
Here we establish the convergence of DP-RSA in the nonconvex setting. Observe that now the cost function (\ref{problem3}) is nonconvex and nonsmooth. One main challenge in the analysis is that we cannot use the optimality gap of function value or iterate for convex functions, nor the stationary condition for nonconvex smooth functions, as the measures of convergence. We use Moreau envelop and proximal point projection \cite{davis2019stochastic} to handle this challenge. The proof can be found in \cite{zhu2022bridging}.

For a continuous weakly convex function $h(\tilde{x})$ with $\tilde{x} \in \mathbb{R}^d$, its Moreau envelope $h_\beta (\tilde{x})$ and proximal point projection $\mathrm{prox}_{\beta h} (\tilde{x})$ are respectively defined as
\begin{align}
    h_\beta (\tilde{x}) & := \mathop {\min}\limits_{\tilde{y} \in \mathbb{R}^d} h(\tilde{y}) + \frac{1}{2\beta} \left \| \tilde{y}-\tilde{x} \right \|^2, \\
    \mathrm{prox}_{\beta h} (\tilde{x}) & := \mathop{\arg \min}\limits_{\tilde{y} \in \mathbb{R}^d} h(\tilde{y}) + \frac{1}{2\beta} \left \| \tilde{y}-\tilde{x} \right \|^2.
\end{align}

Based on the definitions, we immediately have
 \begin{equation}
 \label{eqde}
     \nabla h_\beta (\tilde{x}) = \frac{1}{\beta} (\tilde{x} - \mathrm{prox}_{\beta h} (\tilde{x}) ).
 \end{equation}
More importantly, for any point $\tilde{x} \in \mathbb{R}^d$, its proximal point $\hat{x} = \mathrm{prox}_{\beta h}(\tilde{x})$ satisfies
\begin{align}
    \left \| \hat{x} - \tilde{x} \right \| & = \beta \left \| \nabla h_\beta (\tilde{x}) \right \|, \\
    \|\partial h(\hat{x})\| & \leq \left \| \nabla h_\beta (\tilde{x}) \right \|,
    %h(\hat{x}) & \leq h(\tilde{x}),
\end{align}
where $\partial h(\hat{x})$ is any subgradient of $h(\cdot)$ at $\hat{x}$. That is to say, for a function $h(\cdot)$, a small gradient norm $\left \| \nabla h_\beta (\tilde{x}) \right \|$ implies two facts: $\tilde{x}$ is close to its proximal point $\hat {x}$ and $\hat {x}$ is nearly a stationary point of $h(\cdot)$. Therefore, we can use the measure $\left \| \nabla h_\beta (\tilde{x}) \right \|^2$ to establish the convergence of DP-RSA.

Before proving the convergence of DP-RSA, we further investigate the proposed DP mechanisms. With the Sign-Flipping mechanism, we have
\begin{align}
\E_F [\text{Flip}(\text{sign}(x_0^t - x_k^t)_i)] & = \gamma \text{sign}(x_0^t - x_k^t)_i \notag \\
    & \hspace{-4.2em} + (1-\gamma) \text{sign}(x_k^t - x_0^t)_i,
\end{align}
where $\E_F$ represents the expectation only with respect to the Sign-Flipping operation.
With the Sign-Gaussian mechanism, we have
\begin{align}
\hspace{-1em}    \E_G [\text{Gaussian}&(\text{sign}(x_0^t - x_k^t)_i)] = \left( \Phi(\frac{|(u_k^t)_i|}{\sigma}) \right) \text{sign}(x_0^t - x_k^t)_i \notag \\
    & + \left( 1 - \Phi(\frac{|(u_k^t)_i|}{\sigma}) \right) \text{sign}(x_k^t - x_0^t)_i,
\end{align}
where $\E_G$ represents the expectation only with respect to the Sign-Gaussian operation. To unify the convergence analysis for the two mechanisms, below we let $\tilde{\gamma} = \gamma$ in the Sign-Flipping mechanism, as well as $\gamma = \max_i \Phi(|(u_k^t)_i|/\sigma)$ and $\tilde{\gamma} = \min_i \Phi(|(u_k^t)_i|/\sigma)$ in the Sign-Gaussian mechanism. Thus, we can describe the two mechanisms as
\begin{align}
\E [F/G(\text{sign}(x_0^t - x_k^t)_i)] & \leq \gamma \text{sign}(x_0^t - x_k^t)_i \notag \\
    & \hspace{-4.2em} + (1-\tilde{\gamma}) \text{sign}(x_k^t - x_0^t)_i.
\end{align}

For (\ref{problem3}), define
\begin{align}
h(x) & = \sum\limits_{k \in \mathcal{R}} f_k(x_k) + f_0(x_0) + \sum\limits_{k \in \mathcal{R}} \gamma \lambda \left \| x_k - x_0 \right \|_1.
\end{align}
By the $\rho$-weak convexity of $f_k(x_k)$ and $f_0(x_0)$, $h(x)$ is also $\rho$-weakly convex. %since $\left \| x_k - x_0 \right \|_1$ is convex.
Further, we define $h_{1/\bar{\rho}} (x)$ as the Moreau envelop of $h(x)$ where $\bar{\rho}>0$ is a constant. The following theorem shows that the DP-RSA iterate converges to a neighborhood of a stationary point of $h_{1/\bar{\rho}}(\cdot)$.

\begin{theorem}[Convergence of DP-RSA]
\label{theoremdprsa}
Suppose that Assumptions \ref{a2} and \ref{a3} hold. Set the step size of DP-RSA to $\alpha^t = \alpha = \frac{1}{\sqrt{T}}$. For any constant $\bar{\rho} > \rho$, it holds
\begin{equation}
    \frac{1}{T} \sum\limits_{t=0}^{T-1} \E \left \| \nabla h_{1/\bar{\rho}} (x^t) \right \|^2 \leq \frac{\Delta_1}{\sqrt{T}} + \Delta_2,
\end{equation}
where $\Delta_1, \Delta_2$ are certain constants and $\Delta_2 = O (\lambda^2[b^2 + (1-\tilde{\gamma})^2(r^2+r)])$.
\end{theorem}

According to Theorems \ref{theoremdprsa},
%the malicious messages transmitted by the Byzantine workers lead the iterates to converge to  neighborhoods of a stationary point of $h_{1/\bar{\rho}}(x)$ for DP-RSA and that of $\tilde{h}_{1/\bar{\rho}}(x)$ for RSA.
The learning errors $\Delta_2$ is only linear to $b^2$, the squared number of Byzantine workers, not to the level of Byzantine attacks. If $b=0$, i.e., there are no Byzantine attacks, the learning error $\Delta_2$ is $O(\lambda^2(1 - \tilde{\gamma})(r^2+r))$, which is due to the DP mechanisms. Because the regular workers send back the perturbed signs $F/G(\text{sign} (x_0^t - x_k^t))$, the learning error is relative to the number of regular workers $r$. In the Sign-Flipping mechanism, the larger the probability $1 - \tilde{\gamma}$ of flipping the signs, the larger the learning error. In the Sign-Gaussian mechanism, the larger the variance $\sigma^2$ of added noise, the smaller $\tilde{\gamma}$, and then the larger the learning error. These theoretical observations align with our intuition. Therefore there is a trade-off between privacy preservation and learning performance. If we would like to have a stronger privacy guarantee in the training process, we end up with a larger learning error.

We also observe that the DP mechanisms and the robust stochastic model aggregation rule show an additive influence on the learning performance. Actually the Byzantine attacks and the DP mechanisms have similar impact, such that the DP mechanisms can be regarded as  ``good-will attacks''. Our proposed algorithm can defend against Byzantine attacks, and also has the ability to accommodate the DP mechanisms.

\section{Numerical Experiments}
We provide numerical experiments to verify the effectiveness of DP-RSA on MNIST and CIFAR10 datasets, respectively\footnote{The code is available at https://github.com/oyhah/DP-RSA}. For MNIST, we train a two-layer neural network, each layer containing 50 neurons with tanh activation. In the i.i.d. setting, the 60000 training samples are evenly distributed to 30 workers. In the non-i.i.d. setting, for each digit, half of its samples are evenly distributed to 30 workers, and every 3 workers evenly share the rest half. The regularization term is $f_0(\tilde{x}) = 0.002 \left \| \tilde{x} \right \|^2$. The penalty parameter $\lambda$ is set to 0.01 and the step size $\alpha^t$ is set to be constant as $\alpha = 0.01$.
%\blue{The batch size is 1 thus we use the number of iterations to show the training process.}
For CIFAR10, we train a convolutional neural network (CNN) model with three fully connected layers and two convolutional layers, having about 368,000 parameters in total. In the i.i.d. setting, the 50000 training samples are evenly distributed to 20 workers. In the non-i.i.d. setting, for each class of images, half of its samples are evenly distributed to 20 workers, and every 2 workers evenly share the rest half. The regularization term is $f_0(\tilde{x}) = 0.002 \left \| \tilde{x} \right \|^2$. The penalty parameter $\lambda$ is set to 0.002 and the step size is $\alpha = 0.01$.

%\blue{The batch size is 8 and we use the number of epochs to represent the training process.}

Denote DP-RSA with the Sign-Flipping mechanism and that with the Sign-Gaussian mechanism as DP-RSA(F) and DP-RSA(G), respectively. The privacy loss $\epsilon$ is set to 0.2, 0.4 and 1.38. We consider four baselines: SGD, SignSGD, SGD with Geometric Median (GM), and RSA. All the step sizes are the same as $\alpha=0.01$.
We consider three typical attacks. The first two are applied to the i.i.d. setting and the last one is applied to the non-i.i.d. setting.
(i) Gaussian Attacks: Each Byzantine worker generates a vector where each element is from a Gaussian distribution $\mathcal{N}(0, \sigma_b^2)$, where $\sigma_b = 10000$.
(ii) Sign-Flipping Attacks: Each Byzantine worker calculates the true model and multiplies it with a negative constant $-5$  (see \cite{zhu2022bridging}).
(iii) Sample-Duplicating Attacks: All Byzantine workers pick one regular worker, and duplicate its message.

\vspace{-1em}

\begin{figure}[htb]
    \centering
    \subfigure[Gaussian Attacks]
    {
    \includegraphics[width=0.234\textwidth]{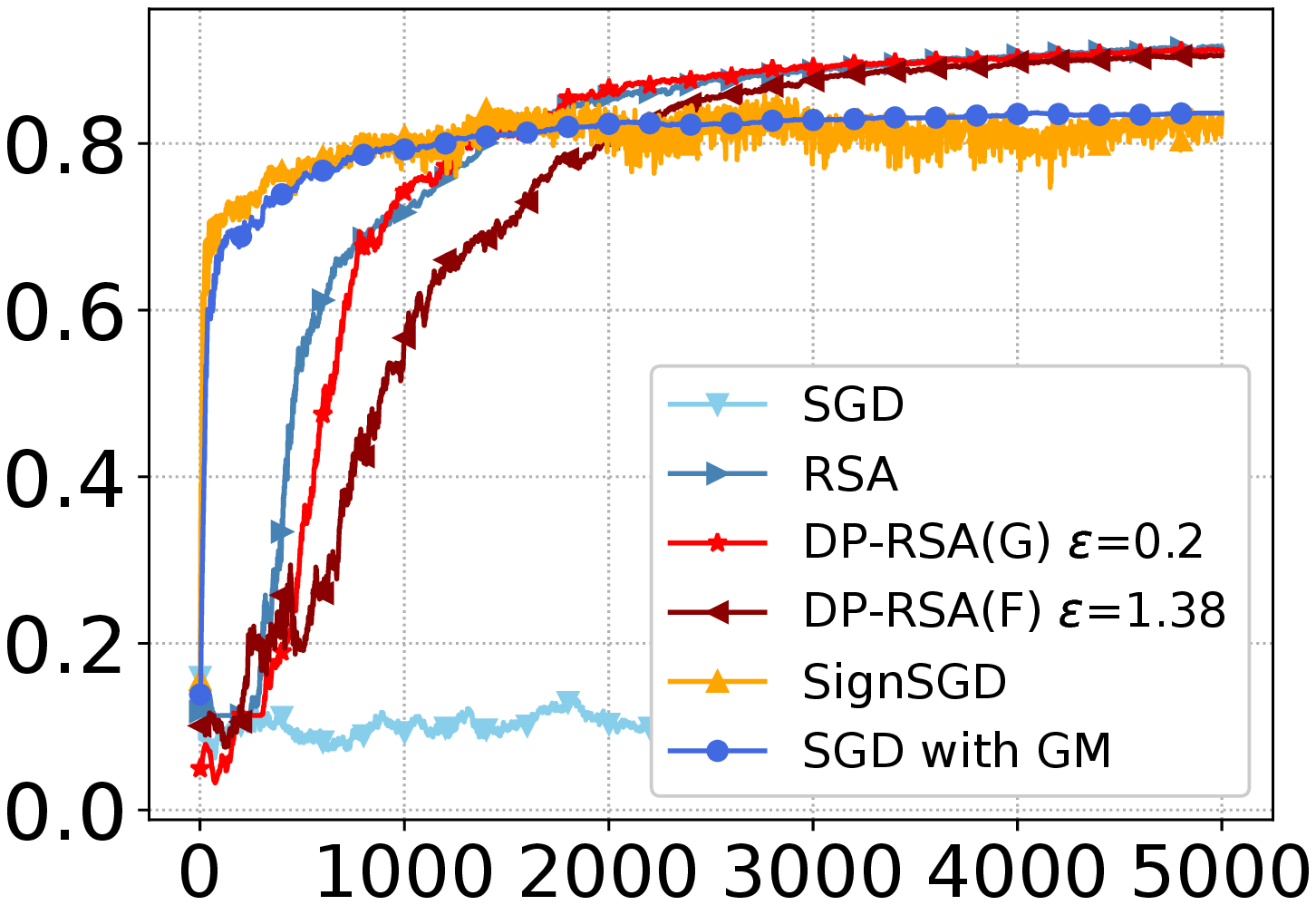}
    }
    \hspace{-1.9em}
    \subfigure[Sample-Duplicating Attacks]
    {
    \includegraphics[width=0.234\textwidth]{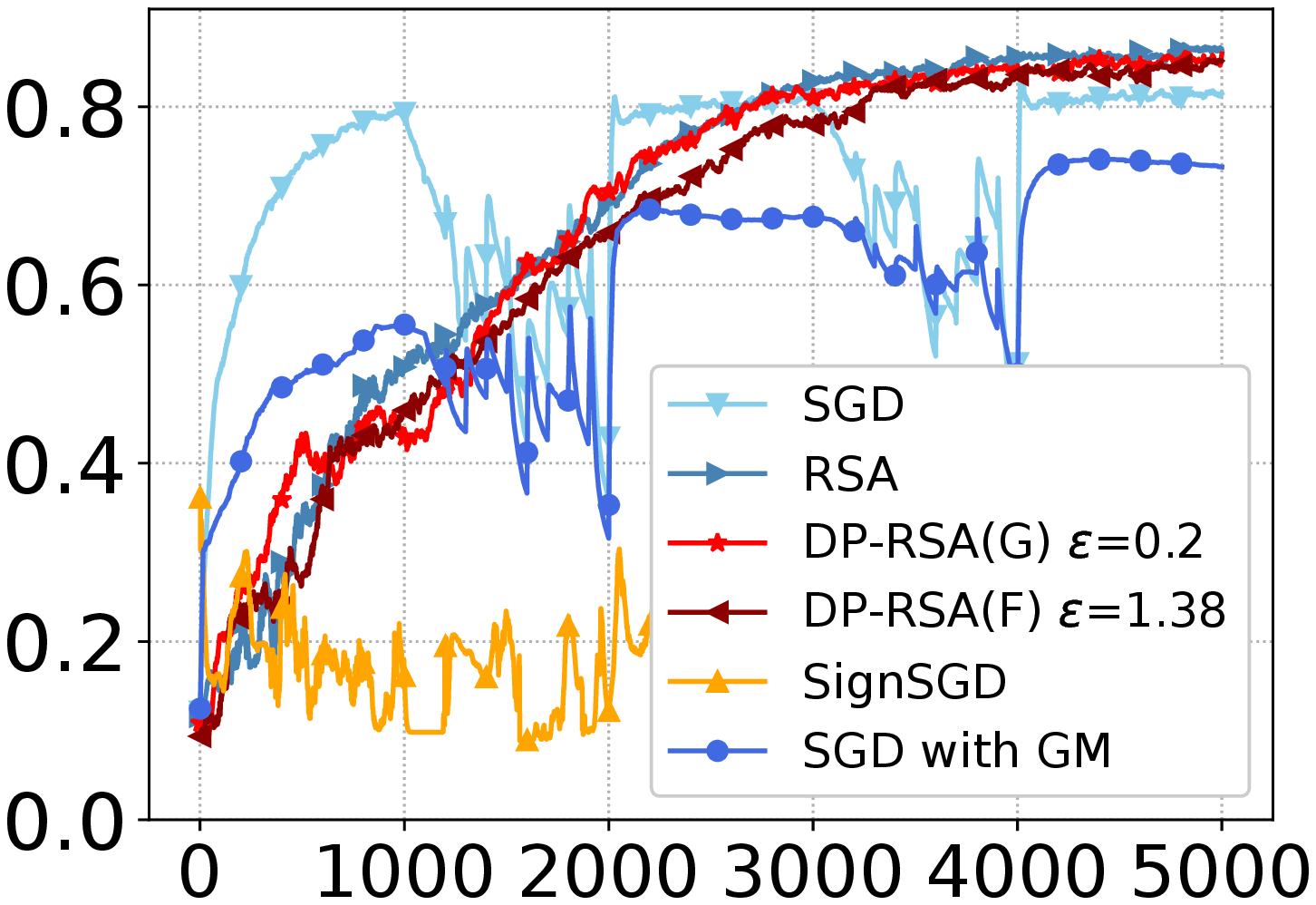}
    } \vspace{-1em}
    \caption{Performance comparisons on MNIST. Horizontal Axis: Number of Iterations. Vertical Axis: Accuracy.}
    \label{ComparisonNN}
\end{figure}
\begin{figure}[htb]
    \centering
    \subfigure[Gaussian Attacks]
    {
    \includegraphics[width=0.234\textwidth]{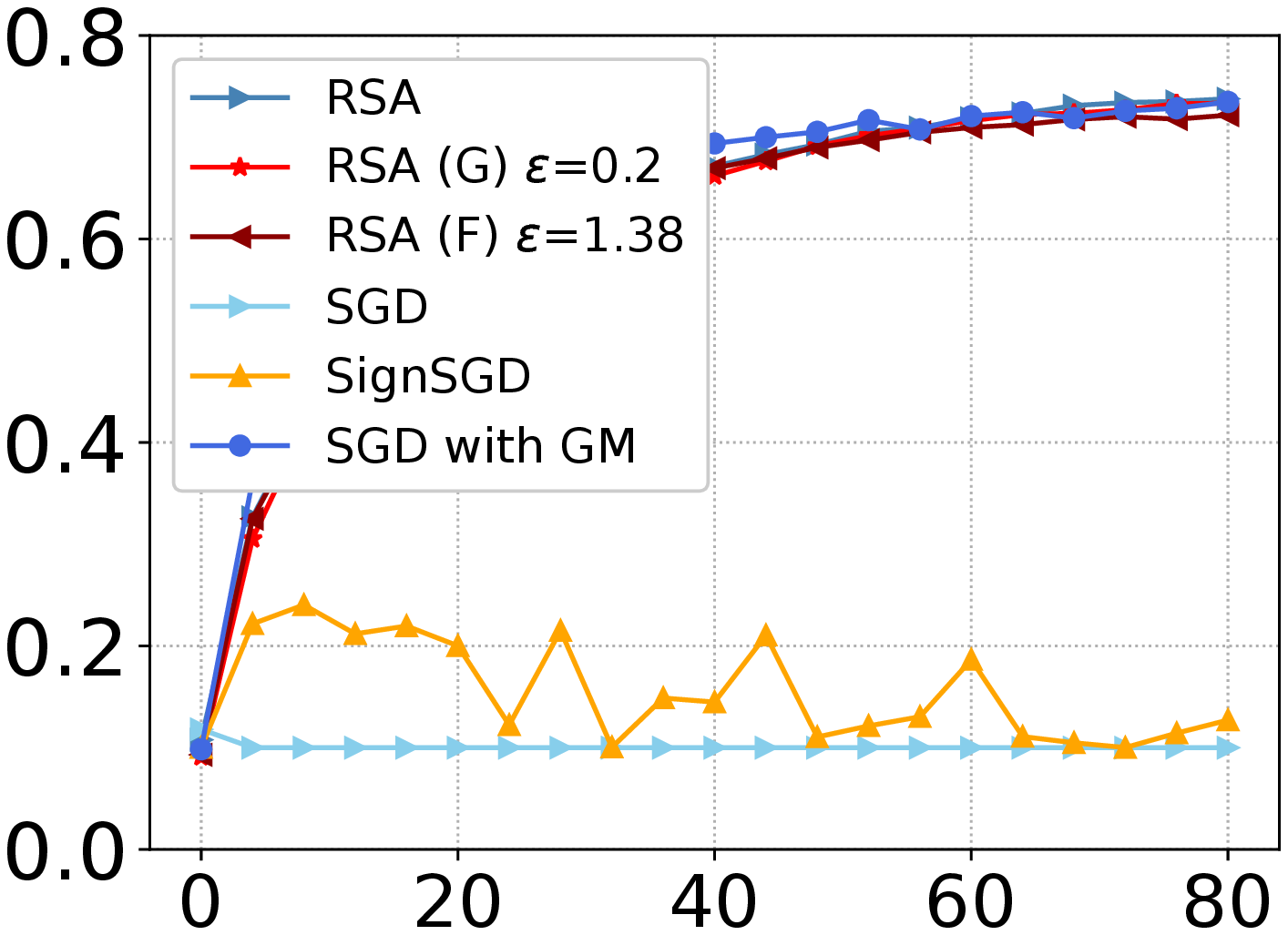}
    }
    \hspace{-1.9em}
    \subfigure[Sample-Duplicating Attacks]
    {
    \includegraphics[width=0.234\textwidth]{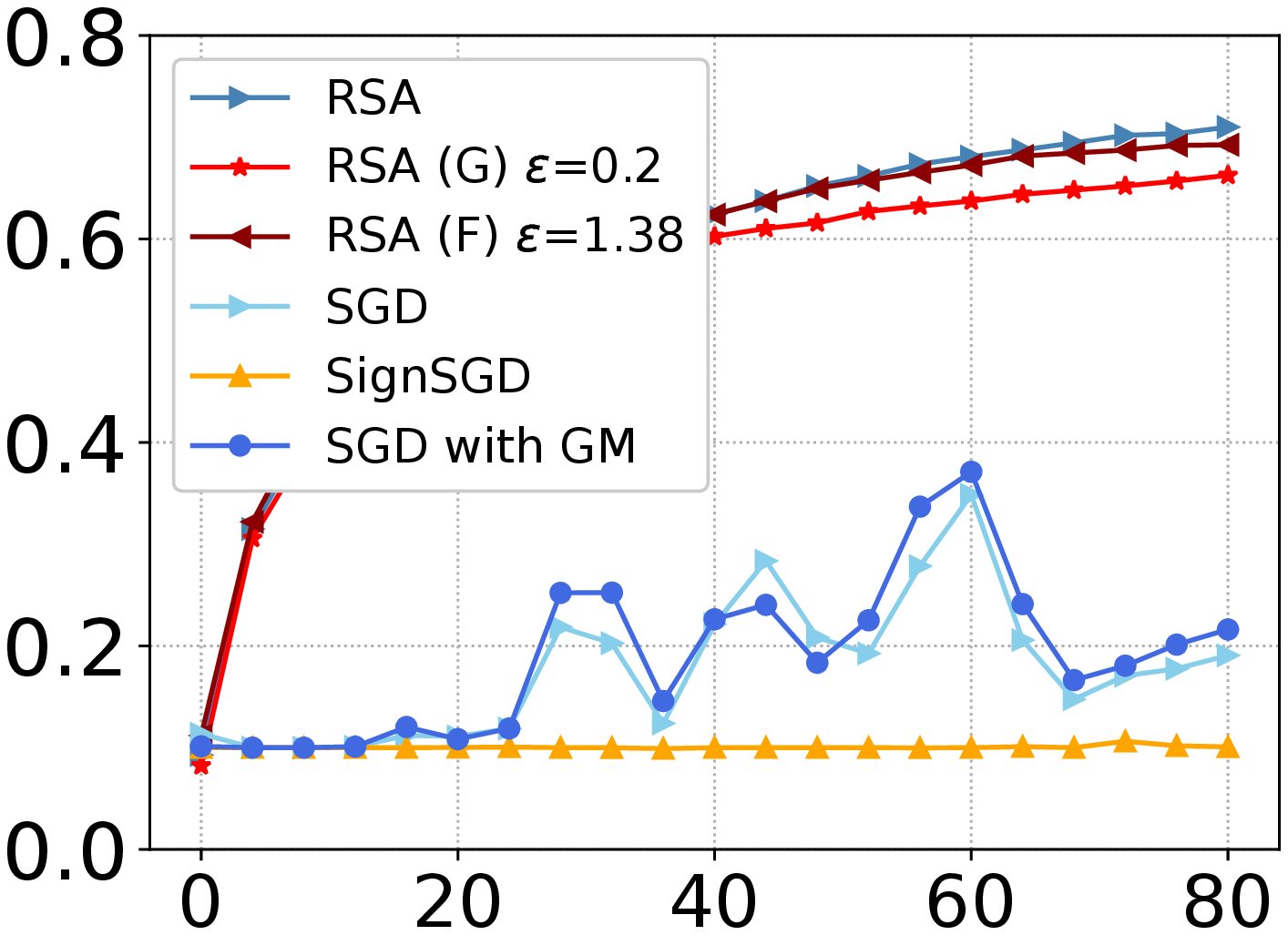}
    } \vspace{-1em}
    \caption{Performance comparisons on CIFAR10. Horizontal Axis: Number of Epochs. Vertical Axis: Accuracy.}
    \label{ComparisonCNN}
\end{figure}

Fig. \ref{ComparisonNN} displays the performance of compared methods on MNIST. Here the number of Byzantine workers is $b=3$. For both attacks, SGD has no defense abilities. DP-RSA performs similarly to RSA, suggesting that the DP mechanisms do not harm the learning performance too much and has no impact on the defense against the Byzantine attacks. Under both attacks, DP-RSA outperforms SignSGD and SGD with GM. For Sample-Duplicating Attacks in the non-i.i.d. setting, SignSGD and SGD with GM is severely influenced by the non-i.i.d. data distribution and almost fail. DP-RSA still stably finds an acceptable solution and is not influenced by the data distribution too much. For the two DP mechanisms used in DP-RSA, Sign-Flipping with large $\epsilon$ has similar performance as Sign-Gaussian with small $\epsilon$, indicating that Sign-Gaussian is more effective in practical applications.

Fig. \ref{ComparisonCNN} shows the performance on CIFAR10. Here the number of Byzantine workers is $b=2$. In the i.i.d. setting, DP-RSA also performs well and is similar to RSA, while SignSGD fails. SGD with GM can defend against Gaussian attack in i.i.d. setting. But in the non-i.i.d. setting, both SignSGD and SGD with GM fail. DP-RSA can successfully defend against the sample-duplicating attack in this case.

In the MNIST experiments, we run 2.5 epochs, and thus the samples are used for nearly 2.5 times. The overall privacy is not too different with the per-epoch privacy. In the CIFAR10 experiments, we run 80 epochs. The analytical moments accountant method can be applied to calculate the overall privacy.  Due to the page limit, we leave more numerical experiments to \cite{zhu2022bridging}. Therein, we show the performance under more attacks, and how the privacy loss $\epsilon$, the number of Byzantine workers $b$, and the penalty parameter $\lambda$ influence the performance of DP-RSA.

\section{Conclusions}

We develop a Byzantine-robust and privacy-preserving federated learning algorithm, DP-RSA, over distributed non-i.i.d. data. The messages transmitted from the workers to the master node are signs of model differences, yielding a communication-efficient implementation. We design two DP mechanisms that provably ensure data privacy. We establish the convergence of DP-RSA for the nononvex cost function and analyze the impact of Byzantine attacks and DP mechanisms on the learning performance. The numerical experiments demonstrate that the proposed DP-RSA can successfully defend against several common Byzantine attacks, for both i.i.d. and non-i.i.d. cases, and protect data privacy without sacrificing the learning performance too much.

% \vspace{-1em}
\section*{Acknowledgements} 
The work of Qing Ling (corresponding author) is supported by National Natural Science Foundation of China under grant 61973324, Guangdong Basic and Applied Basic Research Foundation under grant 2021B1515020094, and Guangdong Provincial Key Laboratory of Computational Science at Sun Yat-Sen University under grant 2020B1212060032.

\bibliography{federated}
\bibliographystyle{named}

\appendix
\onecolumn

% \textbf{Supplementary Materials for ``Bridging Differential Privacy and Byzantine-Robustness via Model Aggregation''}

% \vspace{2em}

\section{Proof of Sign-Flipping Mechanism}

Here we give a simple proof of Theorem \ref{theoremflip}.
\begin{proof}
From the definition of DP we have
\begin{align}
    & \ln \frac{\text{Pr}[\text{Flip}(\text{sign}(x_0^t - x_k^t)) = \text{sign}(x_0^t - x_k^t)]}{\text{Pr}[\text{Flip}(\text{sign}(x_0^t - x_k^t)) = - \text{sign}(x_0^t - x_k^t)]} \\
    = & \ln \frac{\gamma}{1-\gamma}. \notag
\end{align}
This completes the proof.
\end{proof}

\section{Proof of Sign-Gaussian Mechanism}

Here we give the proof of Theorem \ref{theoremgauss}.

\begin{proof}

We measure the privacy loss (PL) as
\begin{align}
    & PL = \ln \frac{P(y_k^t | u_i^k)}{P(y_k^t | u_k^t +v_k^t)} \\
    = & \sum\limits_{i=1}^d \ln \frac{\Phi( (y_k^t)_i (u_k^t)_i / \sigma)}{\Phi( (y_k^t)_i ( (u_k^t)_i + (v_k^t)_i) /\sigma)} \notag \\
    = & \sum\limits_{i=1}^d \ln \Phi( (y_k^t)_i (u_k^t)_i / \sigma) - \ln \Phi( (y_i^k)_i ( (u_k^t)_i + (v_k^t)_i )/\sigma), \notag
\end{align}
where $\Phi (\cdot)$ is the CDF of the standard Gaussian distribution, namely, $\Phi (a) = \frac{1}{\sqrt{2\pi}} \int_{-\infty}^a e^{- t^2/2} dt$.

Denote $q(a) = - \ln \Phi(a)$. Using Taylor expansion at the point $\hat{a}$ yields
\begin{equation}
    q(a) = q(\hat{a}) + q'(\hat{a}) (a - \hat{a}) + \frac{q''(\xi)}{2} (a - \hat{a})^2,
\end{equation}
where $\xi \in (\hat{a}, a)$. It is known that
\begin{equation}
    q'(a) = -\frac{e^{-a^2/2}}{\int_{-\infty}^{a} e^{-s^2/2} ds},
\end{equation}
which is related to Mill's ratio. From \cite{sampford1953some}, we know
\begin{align}
    - \frac{2}{\sqrt{4+a^2} + a}< & q'(a) < a, \\
    0 < & q''(a) < 1.
\end{align}
Thus we can obtain
\begin{equation}
    q(a) < q(\hat{a}) + q'(\hat{a}) (a - \hat{a}) + \frac{1}{2} (a - \hat{a})^2.
\end{equation}

Now let $\hat{a} = (y_k^t)_i (u_k^t)_i/\sigma$ and $a = (y_k^t)_i ((u_k^t)_i + (v_k^t)_i)/ \sigma$. Then, the privacy loss corresponding to the $i$-th dimension can be rewritten as
\begin{align}
    PL_i = & \ln \Phi( (y_k^t)_i (u_k^t)_i / \sigma) - \ln \Phi( (y_i^k)_i ( (u_k^t)_i + (v_k^t)_i )/\sigma)  \\
    < & \ln \Phi( (y_k^t)_i (u_k^t)_i / \sigma) - \ln \Phi( (y_k^t)_i (u_k^t)_i / \sigma) + q'( (y_k^t)_i (u_k^t)_i /\sigma) \frac{(y_k^t)_i (v_k^t)_i}{\sigma} + \frac{1}{2} \frac{(v_k^t)_i^2}{\sigma^2} \notag \\
    = & q'( (y_k^t)_i (u_k^t)_i /\sigma) \frac{(y_k^t)_i (v_k^t)_i}{\sigma} + \frac{1}{2} \frac{(v_k^t)_i^2}{\sigma^2}. \notag
\end{align}

If $y_k^t = 1$, since $q'(a)<0$, we have
\begin{equation}
    PL_i < \frac{1}{2} \frac{(v_k^t)_i^2}{\sigma^2}.
\end{equation}
With $\left \| v_k^t \right \| \leq \Delta u$, we have
\begin{equation}
    PL = \sum\limits_{i=1}^d PL_i < \frac{1}{2} \frac{(\Delta u)^2}{\sigma^2}.
\end{equation}
When $\sigma$ satisfies
\begin{equation}
    \sigma > \frac{\Delta u}{\sqrt{2\epsilon}},
\end{equation}
we can get $PL < \epsilon$.

If $y_k^t = -1$, we have
\begin{align}
    PL_i < & - q'(- (u_k^t)_i/\sigma) \frac{(v_k^t)_i}{\sigma} + \frac{1}{2} \frac{(v_k^t)_i^2}{\sigma^2}  \\
    < & \frac{2}{\sqrt{4 + (u_k^t)_i^2/\sigma^2} - u_k^t/\sigma } \frac{(v_k^t)_i}{\sigma} + \frac{1}{2} \frac{(v_k^t)_i^2}{\sigma^2}. \notag
\end{align}
When $\sigma$ satisfies
\begin{equation}
    \sigma > \frac{2}{3} (u_k^t)_i,
\end{equation}
we have
\begin{equation}
    \sqrt{4 + (u_k^t)_i^2/\sigma^2} - (u_i^k)_i/\sigma > 1.
\end{equation}
Then, we can obtain
\begin{align}
    PL_i < \frac{2 (v_k^t)_i}{\sigma} + \frac{1}{2}\frac{(v_k^t)_i^2}{\sigma^2},
\end{align}
and thus
\begin{equation}
    PL = \sum\limits_{i=1}^d PL_i < \frac{2\sqrt{2}\Delta u}{\sigma} + \frac{1}{2}\frac{(\Delta u)^2}{\sigma^2}.
\end{equation}
To obtain $PL < \epsilon$, we only need to require
\begin{equation}
    \frac{2\sqrt{2}\Delta u}{\sigma} + \frac{1}{2}\frac{(\Delta u)^2}{\sigma^2} < \epsilon.
\end{equation}
To satisfy the above inequality, we need
\begin{equation}
    \sigma > \frac{\Delta u (2\sqrt{2}+ \sqrt{2}\sqrt{4+\epsilon})}{2\epsilon}.
\end{equation}
%Thus only if $\sigma \sim O(\frac{\Delta u}{\sqrt{\epsilon}})$, the above inequality holds.

Further if $\epsilon <8$, which is satisfied in most cases of per-step privacy, we can get
\begin{equation}
    \sigma > \frac{4\Delta u}{{\epsilon}} > \frac{\Delta u (2\sqrt{2}+ \sqrt{2}\sqrt{4+\epsilon})}{2\epsilon}.
\end{equation}
Therefore if $y_i^k = -1$, when $\sigma$ satisfies
\begin{equation}
\label{nega_y}
    \sigma > {\rm max} \{ \max_i \frac{2(u_k^t)_i}{3}, \frac{4\Delta u}{{\epsilon}} \},
\end{equation}
we can get $PL < \epsilon$.

Because $\frac{4\Delta u}{{\epsilon}} > \frac{\Delta u}{\sqrt{2\epsilon}}$ if $\epsilon < 8$, considering both cases of $y_i^k = 1$ and $y_i^k = -1$, we know that when (\ref{nega_y}) is satisfied, the Sign-Gaussian mechanism is $(\epsilon, 0)$-DP.
\end{proof}

\section{Proof of Theorem \ref{theoremdprsa}}

\begin{proof}
First, from Assumption \ref{a3}, we can obtain two useful inequalities
\begin{align}
\label{eqnormsubdp}
    & \E \left \| g_k^t + \lambda F/G(\text{sign}(x_k^t - x_0^t)) \right \|^2  \\
    \leq & 2 \E \left \| g_k^t \right \|^2 + 2\E \left \| \lambda \text{sign}(x_k^t - x_0^t) \right \|^2 \notag \\
    \leq & 2 M^2 + 2 \lambda^2 d, \notag
\end{align}
and
\begin{align}
\label{eqnormsub-1dp}
    & \E \left\|\nabla f_0(x_0^t) + \lambda \sum\limits_{k \in \mathcal{R}} F/G(\text{sign}(x_0^t - x_k^t) + \lambda \sum\limits_{j \in \mathcal{B}} \text{sign}(x_0^t - z_j^t) \right\|^2 \\
    \leq & 2 \E \left \| \nabla f_0(x_0^t) \right \|^2 + 2 \lambda^2 \E \left \| \sum\limits_{k \in \mathcal{R}} F/G(\text{sign}(x_0^t - x_k^t)) + \sum\limits_{j \in \mathcal{B}} \text{sign}(x_0^t - z_j^t) \right \|^2 \notag\\
    \leq & 2 M^2 + 2\lambda^2 K^2 d\notag,
\end{align}
where the second term in the second inequality comes from bounding the $K$ signs.

Further, since $h(\cdot)$ is $\rho$-weakly convex, it holds that
\begin{align}
\label{eqwcdp}
h(y) \geq & h(x) + \left < \partial h(x), y-x \right> - \frac{\rho}{2} \left \| y-x \right \|^2,
\end{align}
where $\partial h(x)$ denotes one subgradient of $h(x)$. We consider a particular subgradient
\begin{align}
\partial h(x)=  [ & \nabla f_{k_1}(x_{k_1}) + \gamma \lambda \text{sign}(x_{k_1} - x_0); \cdots; \\
%    & \cdots; \notag \\
    & \nabla f_{k_R}(x_{k_R}) + \lambda \gamma \text{sign}(x_{k_R} - x_0); \notag \\
    & \nabla f_0(x_0) + \lambda \gamma \sum_{k \in \mathcal{R}} \text{sign}(x_0 - x_k) ] \notag.
\end{align}
Here $k_1, \cdots, k_R$ are indices of the $R$ regular workers.

Let $\hat{x}^t := \mathrm{prox} _{h, 1/\bar{\rho}} (x^t)$. Recalling the definition of Moreau envelop, we have
\begin{align}
\label{eqhdp}
    & \E [h_{1/\bar{\rho}} (x^{t+1}) ]  \\
    \leq & h (\hat{x}^t) + \frac{\bar{\rho}}{2} \E \left \| \hat{x}^t - x^{t+1} \right \|^2 \notag \\
    = & h (\hat{x}^t) + \frac{\bar{\rho}}{2} \E \left \| \hat{x}^t - x^t - (x^{t+1} - x^t) \right \|^2 \notag \\
    = & h (\hat{x}^t) + \frac{\bar{\rho}}{2} \left \| \hat{x}^t - x^t \right \|^2 + \bar{\rho} \E \left< \hat{x}^t - x^t, x^t - x^{t+1} \right> + \frac{\bar{\rho}}{2} \E \left \| x^{t+1} -x^t \right \|^2 \notag \\
    = & h_{ 1/\bar{\rho}} (x^t) + \bar{\rho} \E \left< \hat{x}^t - x^t, x^t - x^{t+1} \right> + \frac{\bar{\rho}}{2} \E \left \| x^{t+1} -x^t \right \|^2 \notag.
\end{align}

Noticing that $x^t=[x_0^t;x_{k_1}^t;\cdots;x_{k_R}^t]$, for the second term at the right-hand side of (\ref{eqhdp}), we can obtain
\begin{align}
\label{eqcross1dp}
    & \E \left< \hat{x}^t - x^t, x^t - x^{t+1} \right> \\
    = & \sum\limits_{k \in \mathcal{R}} \E \left< \hat{x}_k^t - x_k^t, x_k^t - x_k^{t+1} \right> + \E \left< \hat{x}_0^t - x_0^t, x_0^t - x_0^{t+1} \right> \notag \\
    = & \sum\limits_{k \in \mathcal{R}} \E \left< \hat{x}_k^t - x_k^t, \alpha^t \left(g_k^t + \lambda F/G(\text{sign}(x_k^t - x_0^t)) \right) \right> \notag \\
    & + \E \left< \hat{x}_0^t - x_0^t, \alpha^t \left( \nabla f_0(x_0^t) + \lambda \sum\limits_{k \in \mathcal{R}} F/G(\text{sign}(x_0^t - x_k^t)) \right) \right> \notag \\
    & + \left< \hat{x}_0^t - x_0^t, \alpha^t \left( \lambda \sum\limits_{k' \in \mathcal{B}} \text{sign}(x_0^t - z_{k'}^t) \right) \right> \notag \\
    \leq & \sum\limits_{k \in \mathcal{R}} \alpha^t \left< \hat{x}_k^t - x_k^t, \nabla f_k(x_k^t) + \lambda \gamma \text{sign}(x_k^t - x_0^t) \right> + \sum\limits_{k \in \mathcal{R}} \alpha^t \left< \hat{x}_k^t - x_k^t, \lambda (1-\tilde{\gamma}) \text{sign}(x_0^t - x_k^t) \right> \notag \\
    & + \alpha^t \left< \hat{x}_0^t - x_0^t, \nabla f_0(x_0^t) + \lambda \gamma \sum\limits_{k \in \mathcal{R}} \text{sign}(x_0^t - x_k^t)  \right> + \alpha^t \left< \hat{x}_0^t - x_0^t, \lambda (1-\tilde{\gamma}) \sum\limits_{k \in \mathcal{R}} \text{sign}(x_k^t - x_0^t)  \right> \notag \\
    & + \alpha^t \left< \hat{x}_0^t - x_0^t, \lambda \sum\limits_{k' \in \mathcal{B}} \text{sign}(x_0^t - z_{k'}^t) \right> \notag \\
    = & \alpha^t \left < \hat{x}^t-x^t, \partial h(x^t) \right > + \sum\limits_{k \in \mathcal{R}} \alpha^t \left< \hat{x}_k^t - x_k^t, \lambda (1-\tilde{\gamma}) \text{sign}(x_0^t - x_k^t) \right> + \alpha^t \left< \hat{x}_0^t - x_0^t, \lambda (1-\tilde{\gamma}) \sum\limits_{k \in \mathcal{R}} \text{sign}(x_k^t - x_0^t)  \right> \notag \\
    & + \alpha^t \left< \hat{x}_0^t - x_0^t, \lambda \sum\limits_{k' \in \mathcal{B}} \text{sign}(x_0^t - z_{k'}^t) \right> \notag \\
    \leq & - \alpha^t \left (h(x^t) - h(\hat{x}^t) - \frac{\rho}{2}\left \| x^t - \hat{x}^t \right \|^2 \right) + \sum\limits_{k \in \mathcal{R}} \alpha^t \left< \hat{x}_k^t - x_k^t, \lambda (1-\tilde{\gamma}) \text{sign}(x_0^t - x_k^t) \right> \notag \\
    & + \alpha^t \left< \hat{x}_0^t - x_0^t, \lambda (1-\tilde{\gamma}) \sum\limits_{k \in \mathcal{R}} \text{sign}(x_k^t - x_0^t)  \right> + \alpha^t \left< \hat{x}_0^t - x_0^t, \lambda \sum\limits_{k' \in \mathcal{B}} \text{sign}(x_0^t - z_{k'}^t) \right>, \notag
\end{align}
where the second equality is from the updates of $\{x_k\}_{k \in \mathcal{R}}$ and $x_0$, the first equality is from the definition of $\nabla f_k(x_k^t) = \E g_k^t = \E \nabla f_k(x_k, \xi_k^t)$, and the properties of DP functions. And the last inequality is from (\ref{eqwcdp}).

Since $h(\cdot)$ is a $\rho$-weakly convex function and $\bar{\rho} > \rho$, the function $h(x) + \frac{\bar{\rho}}{2} \left \| x^t - x \right \|^2$ is $(\bar{\rho} - \rho)$-strongly convex with respect to $x$. For the first term at the right-hand side of (\ref{eqcross1dp}), we have
\begin{align}
\label{eq-aaaa1dp}
    & h(x^t) - h(\hat{x}^t) - \frac{\rho}{2}\left \| x^t - \hat{x}^t \right \|^2  \\
    = & \left( h(x^t) + \frac{\bar{\rho}}{2} \left \| x^t - x^t \right \|^2 \right) - \left( h(\hat{x}^t) + \frac{\bar{\rho}}{2} \left \| x^t - \hat{x}^t \right \|^2 \right) \notag\\
    &+ \frac{\bar{\rho} - \rho}{2} \left \| x^t - \hat{x}^t \right \|^2 \notag \\
    \geq & (\bar{\rho} - \rho) \left \| x^t - \hat{x}^t \right \|^2, \notag
\end{align}
where the second equality is from (\ref{eqde}).

For the second term at the right-hand side of (\ref{eqcross1dp}), it holds for any $\xi > 0$ that
\begin{align}
\label{eqhkdp}
    & \sum\limits_{k \in \mathcal{R}} \E \left< \hat{x}_k^t - x_k^t, (1-\tilde{\gamma}) \lambda  \text{sign}(x_0^t - x_k^t) \right>  \\
    \leq & \sum\limits_{k \in \mathcal{R}} \xi \E \left \| \hat{x}_k^t - x_k^t \right \|^2 + \sum\limits_{k \in \mathcal{R}} \frac{\lambda^2(1-\tilde{\gamma})^2}{\xi} \E \left \| \text{sign}(x_0^t - x_k^t) \right \|^2 \notag \\
    \leq & \sum\limits_{k \in \mathcal{R}} \xi \E \left \| \hat{x}_k^t - x_k^t \right \|^2 + \frac{\lambda^2(1-\tilde{\gamma})^2rd}{\xi}. \notag
\end{align}

For the third term at the right-hand side of (\ref{eqcross1dp}), it holds for the same $\xi > 0$ that
\begin{align}
\label{eqh0dp}
    & \E \left< \hat{x}_0^t - x_0^t, (1-\tilde{\gamma}) \lambda \sum\limits_{k \in \mathcal{R}} \text{sign}(x_k^t - x_0^t) \right>  \\
    \leq & \xi \E \left \| \hat{x}_0^t - x_0^t \right \|^2 + \frac{\lambda^2(1-\tilde{\gamma})^2}{\xi} \E \left \| \sum\limits_{k \in \mathcal{R}} \text{sign}(x_k^t - x_0^t) \right \|^2 \notag \\
    \leq & \xi \E \left \| \hat{x}_0^t - x_0^t \right \|^2 + \frac{\lambda^2(1-\tilde{\gamma})^2r^2d}{\xi}. \notag
\end{align}

For the fourth term at the right-hand side of (\ref{eqcross1dp}), it holds for any $\eta > 0$ that
\begin{align}
\label{eqbydp}
    & \left< \hat{x}_0^t - x_0^t, \lambda \sum\limits_{k' \in \mathcal{B}} \text{sign}(x_0^t - z_{k'}^t) \right>  \\
    \leq & \eta  \left \| \hat{x}_0^t - x_0^t \right \|^2 + \frac{\lambda^2}{\eta}  \left \| \sum\limits_{k' \in \mathcal{B}} \text{sign}(x_0^t - z_{k'}^t) \right \|^2 \notag \\
    \leq & \eta \left \| \hat{x}_0^t - x_0^t \right \|^2 + \frac{\lambda^2b^2d}{\eta}. \notag
\end{align}

With \eqref{eq-aaaa1dp} and \eqref{eqhkdp}, \eqref{eqh0dp}, \eqref{eqbydp}, for (\ref{eqcross1dp}) we can obtain
\begin{align}
\label{eqcrossdp}
    & \E \left< \hat{x}^t - x^t, x^t - x^{t+1} \right> \\
    \leq & - \alpha^t (\bar{\rho} - \rho) \left \| x^t - \hat{x}^t \right \|^2 + \sum\limits_{k \in \mathcal{R}} \alpha^t \xi \E \left \| \hat{x}_k^t - x_k^t \right \|^2 + \alpha^t \xi \E \left \| \hat{x}_0^t - x_0^t \right \|^2 + \alpha^t \eta \left \| \hat{x}_0^t - x_0^t \right \|^2 \notag \\
    & + \frac{\alpha^t\lambda^2(1-\tilde{\gamma})^2(r^2+r)d}{\xi} + \frac{\alpha^t\lambda^2b^2d}{\eta} \notag \\
    = & - \alpha^t (\bar{\rho} - \rho - \xi) \sum\limits_{k \in \mathcal{R}}\left \| x_k^t - \hat{x}_k^t \right \|^2 - \alpha^t (\bar{\rho} - \rho - \xi - \eta) \left \| x_0^t - \hat{x}_0^t \right \|^2 + \frac{\alpha^t\lambda^2(1-\tilde{\gamma})^2(r^2+r)d}{\xi} + \frac{\alpha^t\lambda^2b^2d}{\eta} \notag \\
    \leq & - \alpha^t (\bar{\rho} - \rho - \xi - \eta) \left \| \hat{x}^t - x^t \right \|^2 + \frac{\alpha^t\lambda^2(1-\tilde{\gamma})^2(r^2+r)d}{\xi} + \frac{\alpha^t\lambda^2b^2d}{\eta} \notag \\
    = & - \frac{\alpha^t (\bar{\rho} - \rho - \xi - \eta)}{\bar{\rho}^2} \left \| \nabla h_{1/\bar{\rho}} (x^t) \right \|^2 + \frac{\alpha^t\lambda^2(1-\tilde{\gamma})^2(r^2+r)d}{\xi} + \frac{\alpha^t\lambda^2b^2d}{\eta}, \notag
\end{align}
where the last equality is from (\ref{eqde}).

For the third term at the right-hand side of (\ref{eqhdp}), we have
\begin{align}
\label{eqsquaredp}
    & \E \left \| x^{t+1} -x^t \right \|^2 \\
    = & \sum\limits_{k \in \mathcal{R}} \E \left \| x_k^{t+1} -x_k^t \right \|^2 + \E \left \| x_0^{t+1} -x_0^t \right \|^2 \notag \\
    = & \sum\limits_{k \in \mathcal{R}} (\alpha^t)^2 \E \left \| g_k^t + \lambda F/G(\text{sign}(x_k^t - x_0^t)) \right \|^2 + (\alpha^t)^2 \left\|\nabla f_0(x_0^t) + \lambda \sum\limits_{k \in \mathcal{R}} F/G(\text{sign}(x_0^t - x_k^t)) + \lambda \sum\limits_{k' \in \mathcal{B}} \text{sign}(x_0^t - z_{k'}^t) \right\|^2 \notag \\
    \leq & (\alpha^t)^2 r (2M^2 + 2 \lambda^2d) + (\alpha^t)^2 (2M^2 + 2\lambda^2K^2d) \notag \\
    = & 2 (\alpha^t)^2 \left( (r+1)M^2 + (r+K^2)\lambda^2d \right), \notag
\end{align}
where the inequality is from (\ref{eqnormsubdp}) and (\ref{eqnormsub-1dp}).

Substituting (\ref{eqcrossdp}), (\ref{eqsquaredp}) into (\ref{eqhdp}), we can obtain
\begin{align}
    & \E [h_{1/\bar{\rho}} (x^{t+1}) ]   \\
    \leq & \E [h_{1/\bar{\rho}} (x^t) ] - \E \frac{\alpha^t(\bar{\rho} - \rho - \eta - \xi)}{\bar{\rho}} \left \| \nabla h_{1/\bar{\rho}} (x^t) \right \|^2 + (\alpha^t)^2 \bar{\rho} \left [ (r+1)M^2 + (r + K^2)\lambda^2d \right] \notag \\
    & + \alpha^t \bar{\rho} \lambda^2 d\left( \frac{b^2}{\eta} + \frac{(1-\tilde{\gamma})^2(r^2+r)}{\xi} \right). \notag
\end{align}
Therefore we can obtain
\begin{align}
    & \E \left \| \nabla h_{1/\bar{\rho}} (x^t) \right \|^2  \\
    \leq &  \frac{\bar{\rho}}{\alpha^t(\bar{\rho} - \rho - \eta - \xi)} \left( \E [h_{1/\bar{\rho}} (x^t) ] - \E [h_{1/\bar{\rho}} (x^{t+1}) ] \right) \notag \\
    & + \alpha^t \frac{\bar{\rho}^2}{\bar{\rho} - \rho - \eta - \xi} \left [ (r+1)M^2 + (r + K^2)\lambda^2d \right] +  \frac{\bar{\rho}^2\lambda^2d}{\bar{\rho} - \rho - \eta - \xi} \left( \frac{b^2}{\eta} + \frac{(1-\tilde{\gamma})^2(r^2+r)}{\xi} \right). \notag
\end{align}

Here we use a constant step size such that $\alpha^t = \alpha$ for all $t$. Letting $h_m = \min h_{1/\bar{\rho}} (x)$ and applying telescopic cancellation through $t=0$ to $t=T-1$, we have
\begin{align}
    & \frac{1}{T} \sum\limits_{t=0}^{T-1} \E \left \| \nabla h_{1/\bar{\rho}} (x^t) \right \|^2  \\
    \leq & \frac{\bar{\rho}}{T\alpha(\bar{\rho} - \rho - \eta - \xi)} \left( [h_{1/\bar{\rho}} (x^0) ] - \E [h_{1/\bar{\rho}} (x^{T-1}) ] \right) \notag \\
    & + \alpha \frac{\bar{\rho}^2}{\bar{\rho} - \rho - \eta - \xi} \left [ (r+1)M^2 + (r + K^2)\lambda^2d \right] +  \frac{\bar{\rho}^2\lambda^2d}{\bar{\rho} - \rho - \eta - \xi} \left( \frac{b^2}{\eta} + \frac{(1-\tilde{\gamma})^2(r^2+r)}{\xi} \right) \notag \\
    \leq & \frac{\bar{\rho}}{T\alpha(\bar{\rho} - \rho - \eta -\xi)} \left( [h_{1/\bar{\rho}} (x^0) ] - h_m ] \right) \notag \\
    & + \alpha \frac{\bar{\rho}^2}{\bar{\rho} - \rho - \eta - \xi} \left [ (r+1)M^2 + (r + K^2)\lambda^2d \right] +  \frac{\bar{\rho}^2\lambda^2d}{\bar{\rho} - \rho - \eta - \xi} \left( \frac{b^2}{\eta} + \frac{(1-\tilde{\gamma})^2(r^2+r)}{\xi} \right). \notag
\end{align}

If the step size is chosen as $\alpha = \frac{1}{\sqrt{T}}$, we finally get
\begin{align}
    \frac{1}{T} \sum\limits_{t=0}^{T-1} \E \left \| \nabla h_{1/\bar{\rho}} (x^t) \right \|^2 \leq \frac{\Delta_1}{\sqrt{T}} + \Delta_2,
\end{align}
where
\begin{align}
    \Delta_1 &:= \frac{\bar{\rho}}{(\bar{\rho} - \rho - \eta - \xi)} \left( h_{1/\bar{\rho}} (x^0) - h_m ] \right) + \frac{\bar{\rho}^2}{\bar{\rho} - \rho - \eta - \xi} \left [ (r+1)M^2 + (r + K^2)\lambda^2d \right], \\
    \Delta_2 &:= \frac{\bar{\rho}^2\lambda^2d}{\bar{\rho} - \rho - \eta - \xi} \left( \frac{b^2}{\eta} + \frac{(1-\tilde{\gamma})^2(r^2+r)}{\xi} \right).
\end{align}
Note that we need to choose $\eta$ and $\xi$ such that $\bar{\rho} - \rho - \eta - \xi > 0$, for example, $\eta=\xi=\frac{\bar{\rho} - \rho}{3}$.
\end{proof}

\section{Additional Numerical Experiments}

In this part we provide more numerical experiments to demonstrate the effectiveness of DP-RSA.

\subsection{Experiment Setting of Compared Methods}

For RSA, the step size, regularization parameter and penalty parameter $\lambda$ are the same as those in DP-RSA in MNIST and CIFAR10 experiments, respectively.
For SGD, SignSGD, SGD with GM, RSA, DP-RSA, the step sizes are all given by $\alpha^t = \alpha = 0.01$. SignSGD transmits the element-wise signs of stochastic gradients at all the workers to the master node. SGD with GM applies geometric median aggregation of the transmitted stochastic gradients from all the workers at the master node.
In MNIST, the batch size is 1 and we use the number of iterations to show the training process.
In CIFAR10, the batch size is 8 and we use the number of epochs to represent the training process.

\subsection{Impact of Privacy Loss $\epsilon$}

\begin{figure}[htb]
    \centering
    \subfigure[Gaussian Attacks]{
    \includegraphics[width=0.30\textwidth]{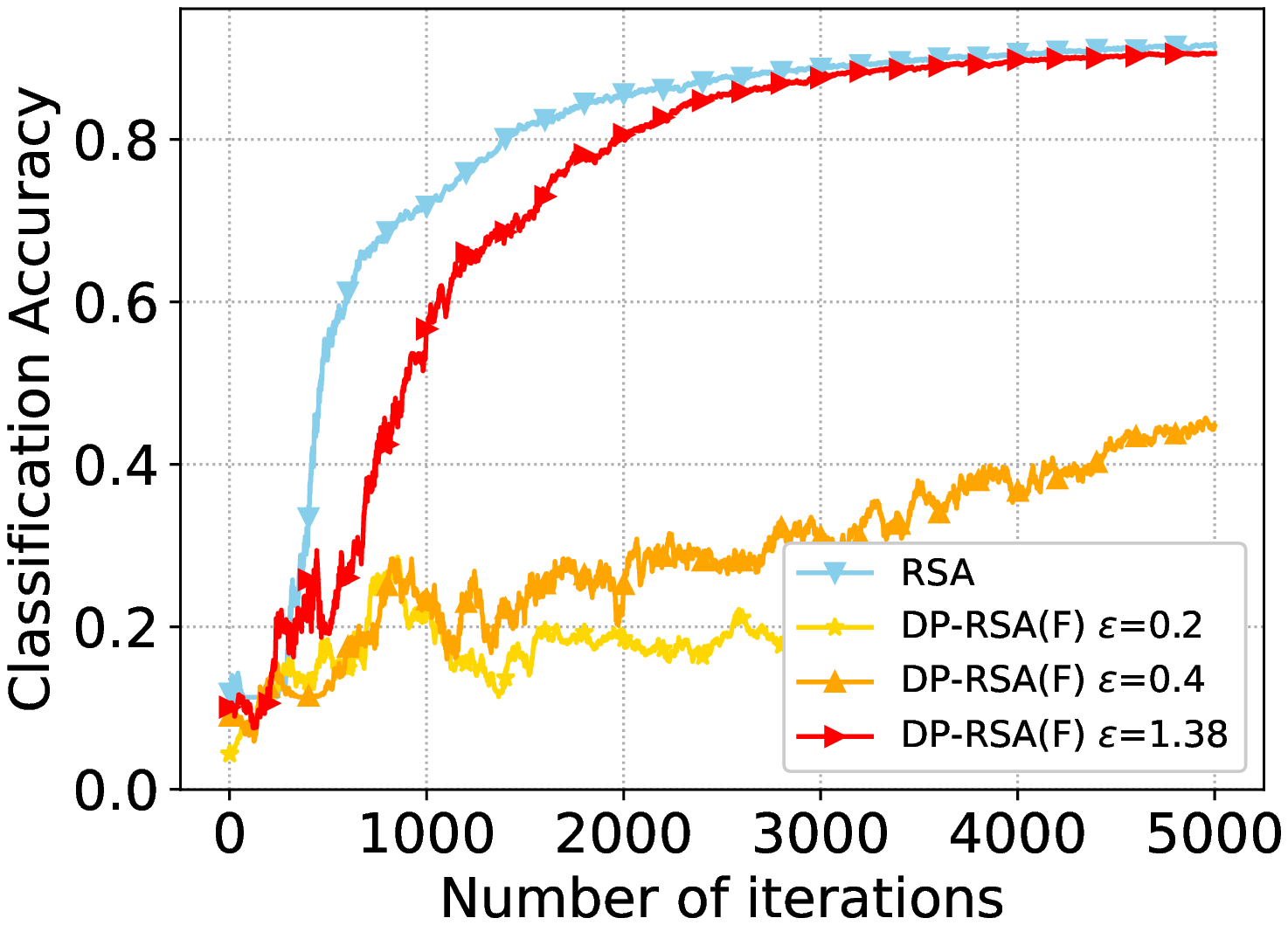}
    }
    \subfigure[Sign-Flipping Attacks]{
    \includegraphics[width=0.30\textwidth]{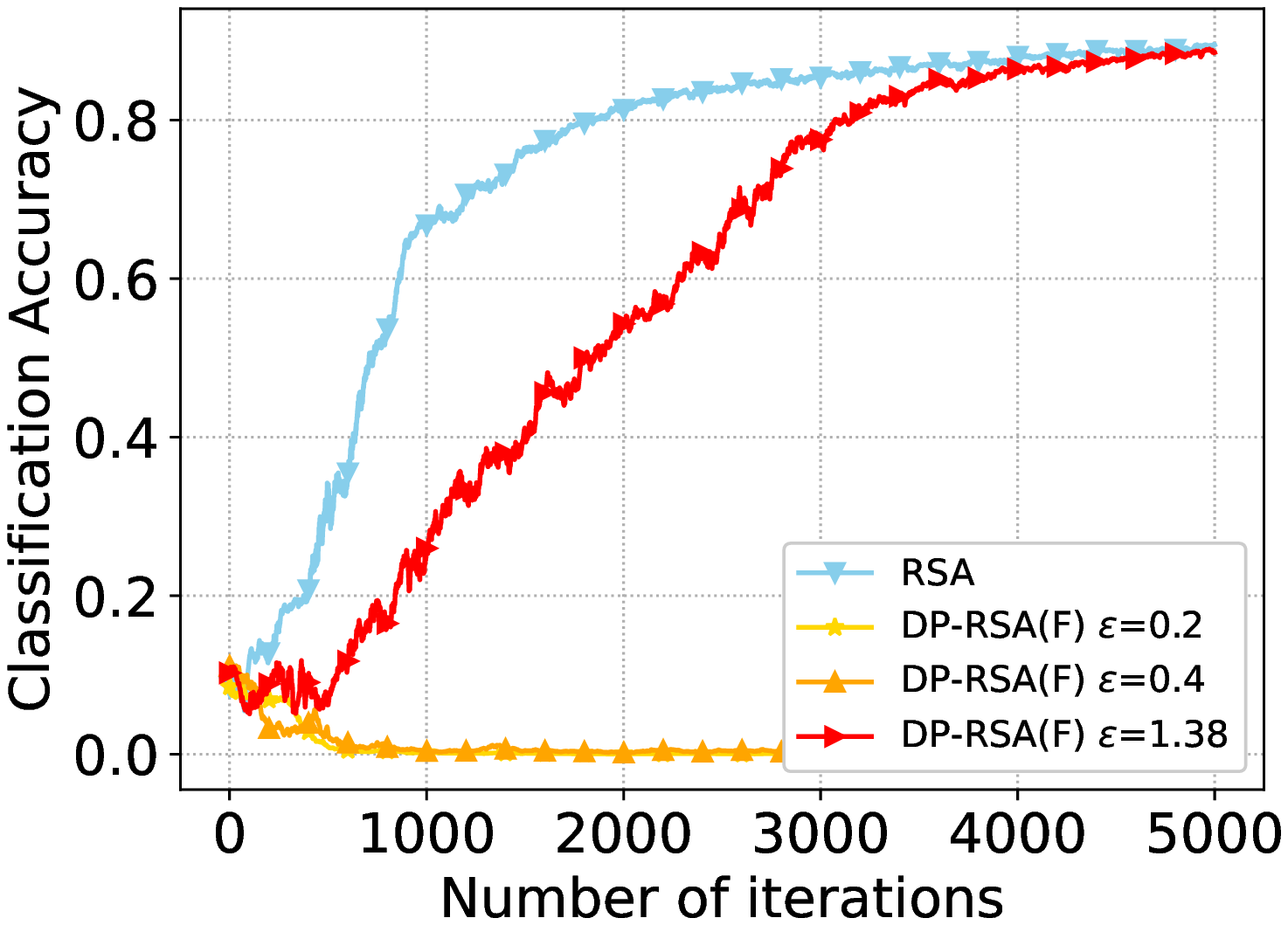}
    }
    \subfigure[Sample-Duplicating Attacks]{
    \includegraphics[width=0.30\textwidth]{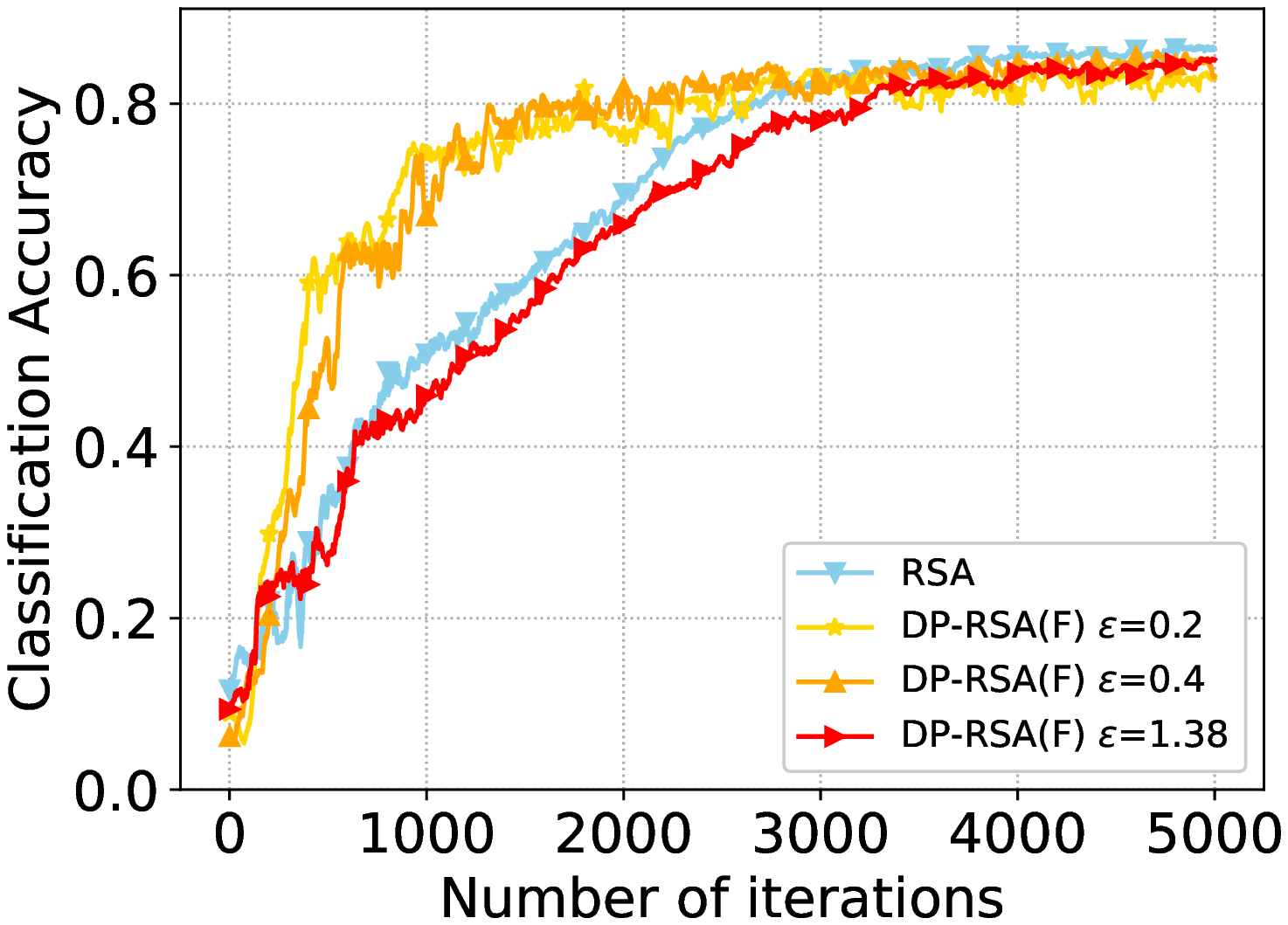}
    }

    \caption{Impact of privacy loss $\epsilon$ on Sign-Flipping Mechanism over MNIST dataset.}
    \label{epsilon-mnist-flip}
\end{figure}

\begin{figure}[htb]
    \centering
    \subfigure[Gaussian Attacks]{
    \includegraphics[width=0.30\textwidth]{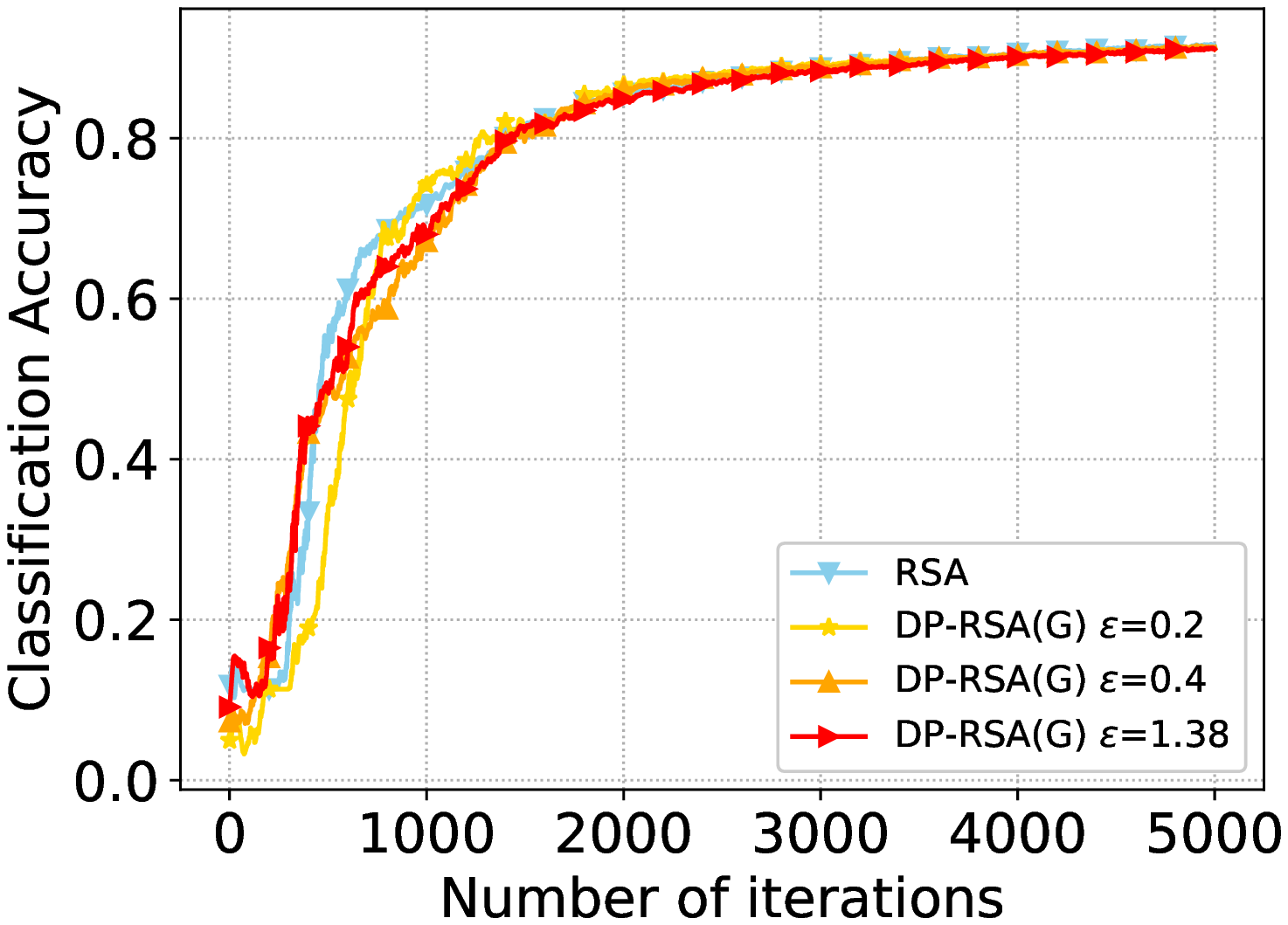}
    }
    \subfigure[Sign-Flipping Attacks]{
    \includegraphics[width=0.30\textwidth]{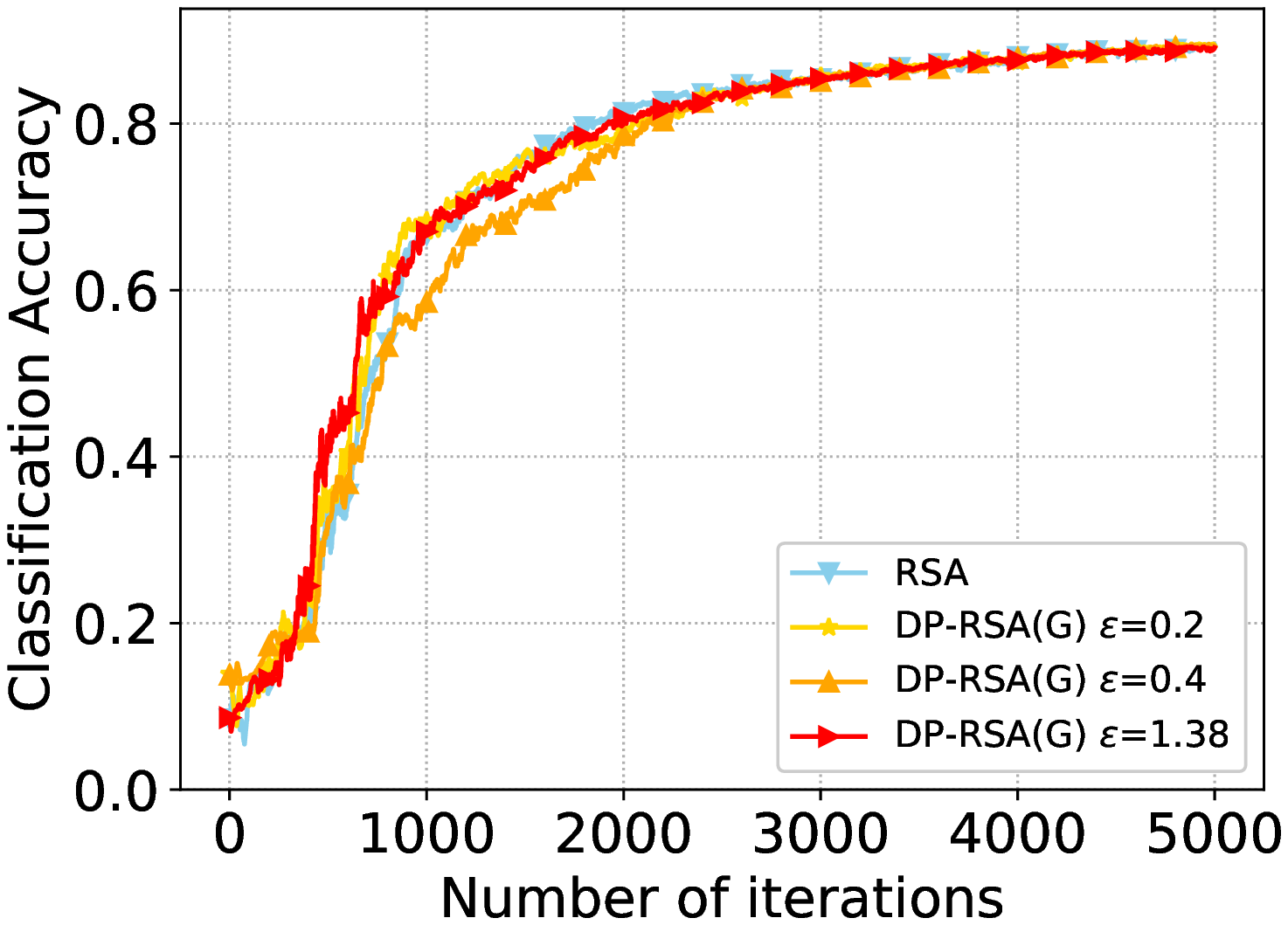}
    }
    \subfigure[Sample-Duplicating Attacks]{
    \includegraphics[width=0.30\textwidth]{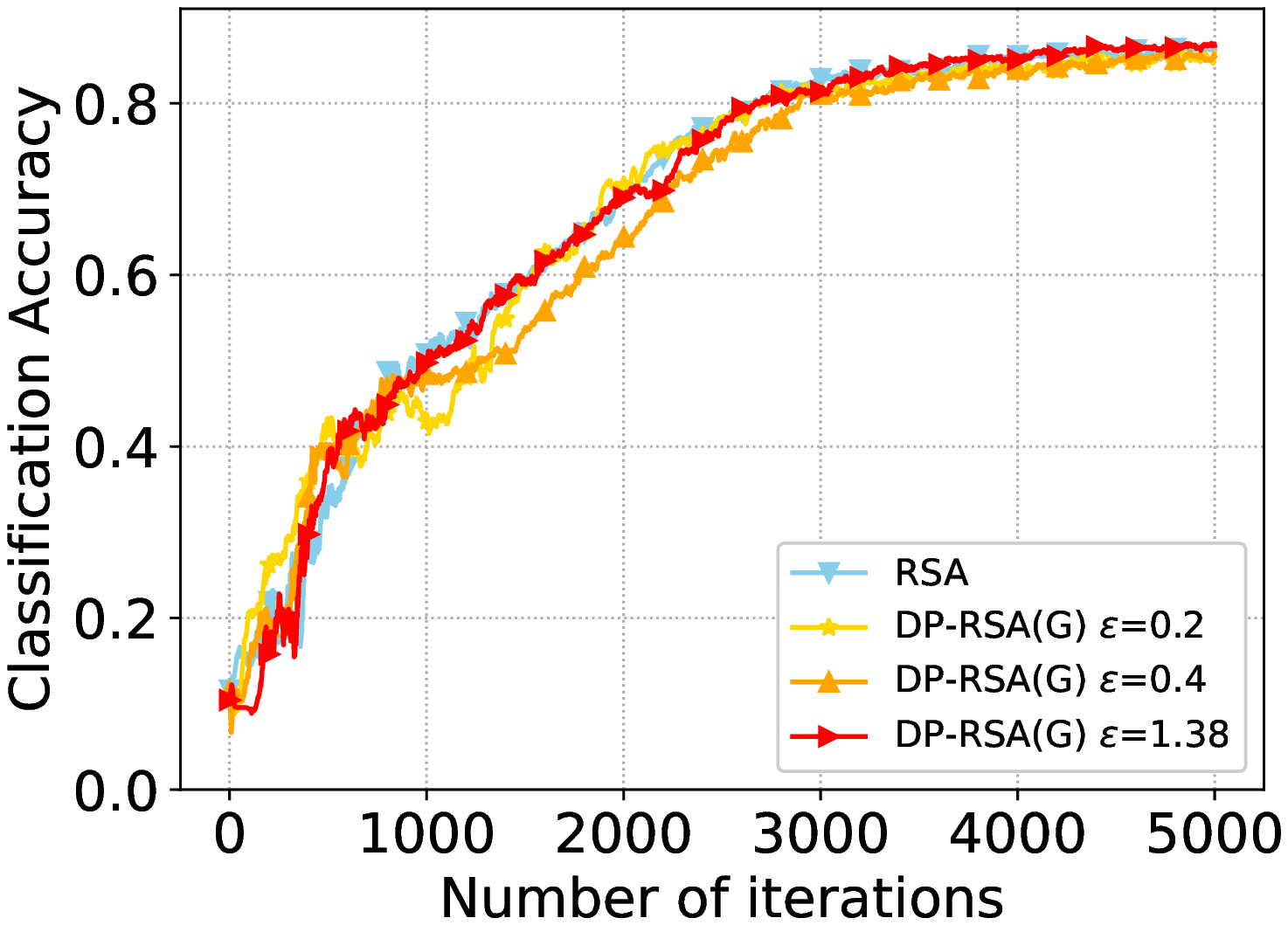}
    }
    \caption{Impact of privacy loss $\epsilon$ on Sign-Gaussian Mechanism over MNIST dataset.}
    \label{epsilon-mnist-gauss}
\end{figure}

\begin{figure}[htb]
    \centering
    \subfigure[Gaussian Attacks]{
    \includegraphics[width=0.30\textwidth]{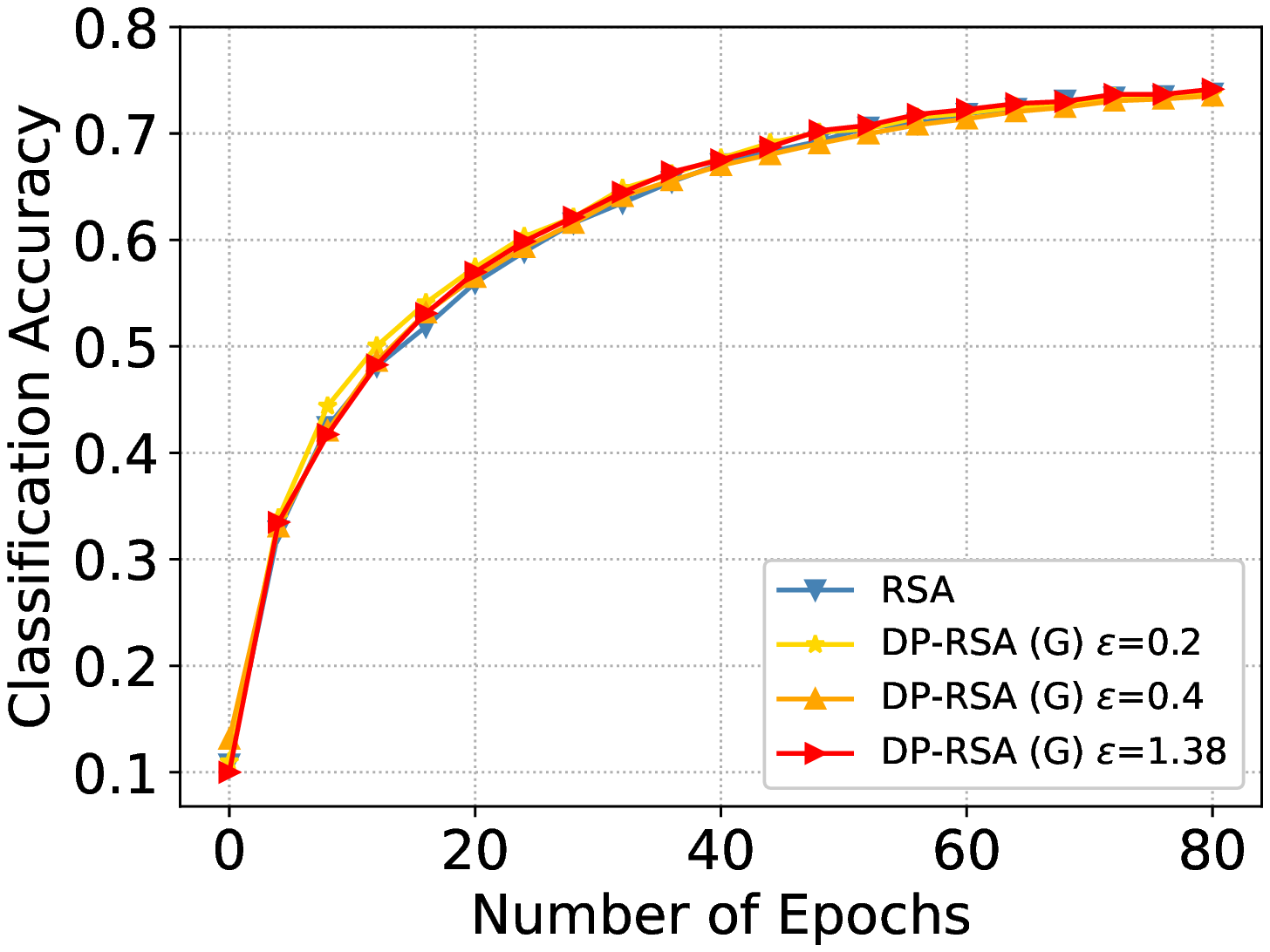}
    }
    \subfigure[Sign-Flipping Attacks]{
    \includegraphics[width=0.30\textwidth]{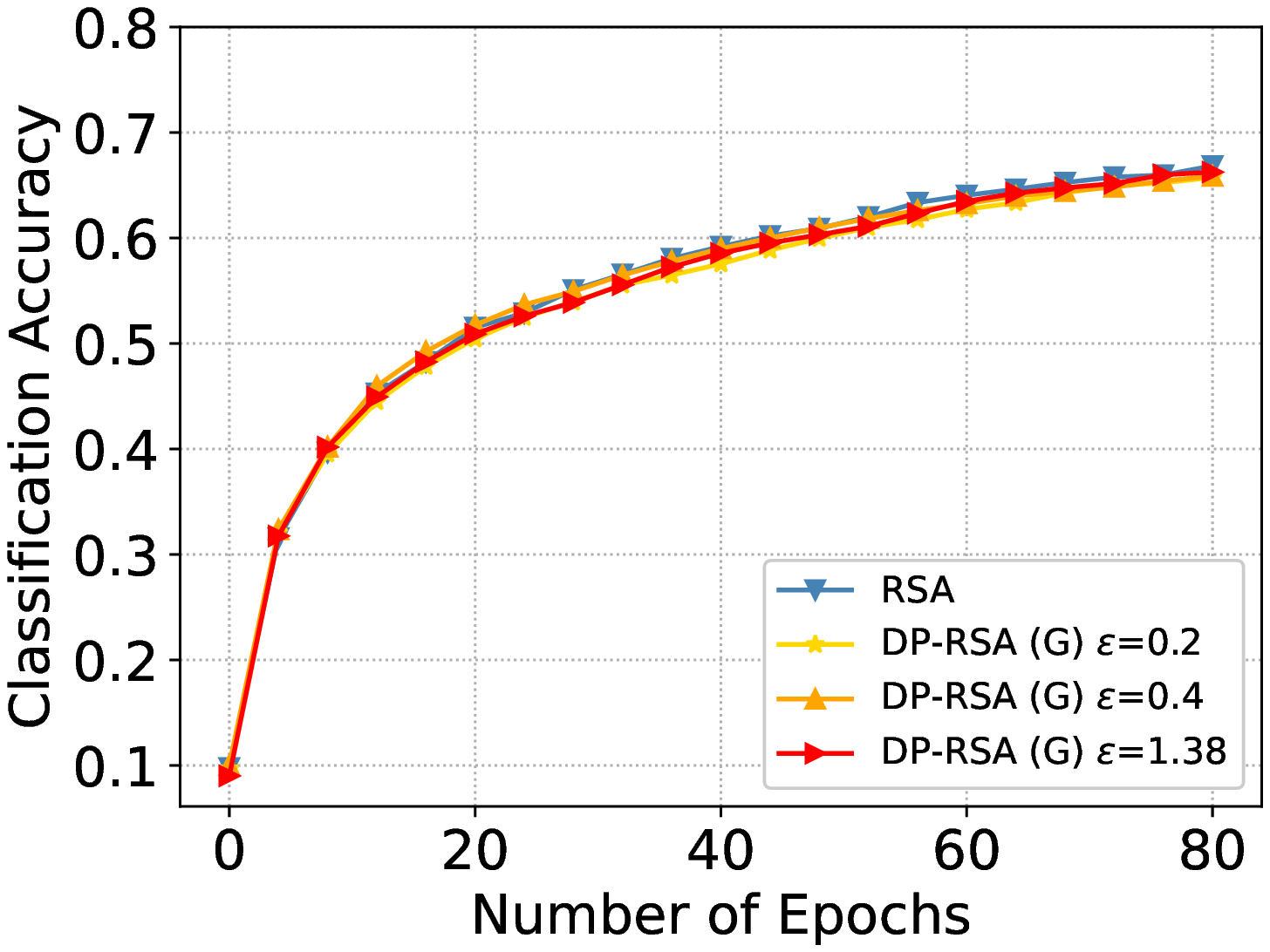}
    }
    \subfigure[Sample-Duplicating Attacks]{
    \includegraphics[width=0.30\textwidth]{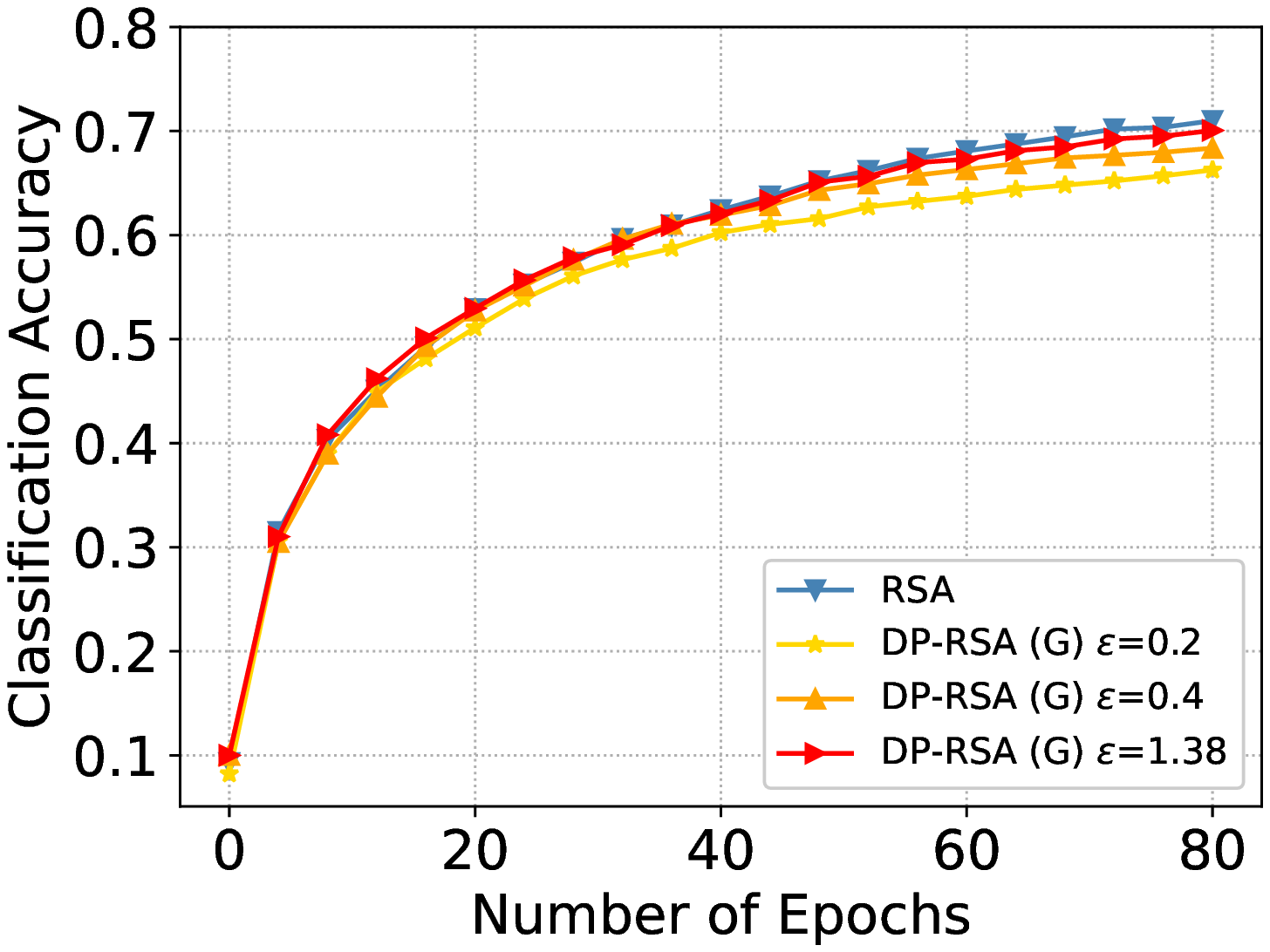}
    }
    \caption{Impact of privacy loss $\epsilon$ on Sign-Gaussian Mechanism over CIFAR10 dataset.}
    \label{epsilon-cifar}
\end{figure}

Here we investigate the impact of privacy loss $\epsilon$ on the learning performance. Other experimental settings are the same as those in the main text.

Figs. \ref{epsilon-mnist-flip} and \ref{epsilon-mnist-gauss} show the impact of $\epsilon$ on the Sign-Flipping and Sign-Gaussian mechanisms over the MNIST dataset, respectively. We consider Gaussian Attacks, Sign-Flipping Attacks in the i.i.d. setting and Sample-Duplicating Attacks in the non-i.i.d. setting. For the Sign-Flipping mechanism, to achieve the same privacy loss as the Sign-Gaussian mechanism, the probability of flipping the signs should be large, which hurts the convergence especially for the two attacks in the i.i.d. case. Thus, the Sign-Flipping mechanism may be severely influenced by the targeted privacy level. For example, to reach a privacy loss $\epsilon$ of 1.38, we should flip the signs with probability 0.2. When $\epsilon$ is 0.2, the corresponding probability of flipping the signs increases to 0.45.
%and DP-RSA with the Sign-Flipping mechanism are able to converge and defend against the attacks.
Although Sign-Flipping is a straightforward mechanism, its ability to preserve data privacy is not satisfactory when we expect a small learning error. However, the Sign-Gaussian is less influenced by the privacy level. As the training process evolves, the difference between local and global models gradually decreases. Thus, the added noise becomes smaller, and eventually, very small noise can guarantee data privacy.

Fig. \ref{epsilon-cifar} demonstrates the impact of $\epsilon$ on the Sign-Gaussian mechanism over the CIFAR10 dataset. The performance is also less influenced by the privacy level.

\subsection{Impact of Number of Byzantine Workers $b$}

\begin{figure}[htb]
    \centering
    \subfigure[MNIST Dataset]{
    \includegraphics[width=0.30\textwidth]{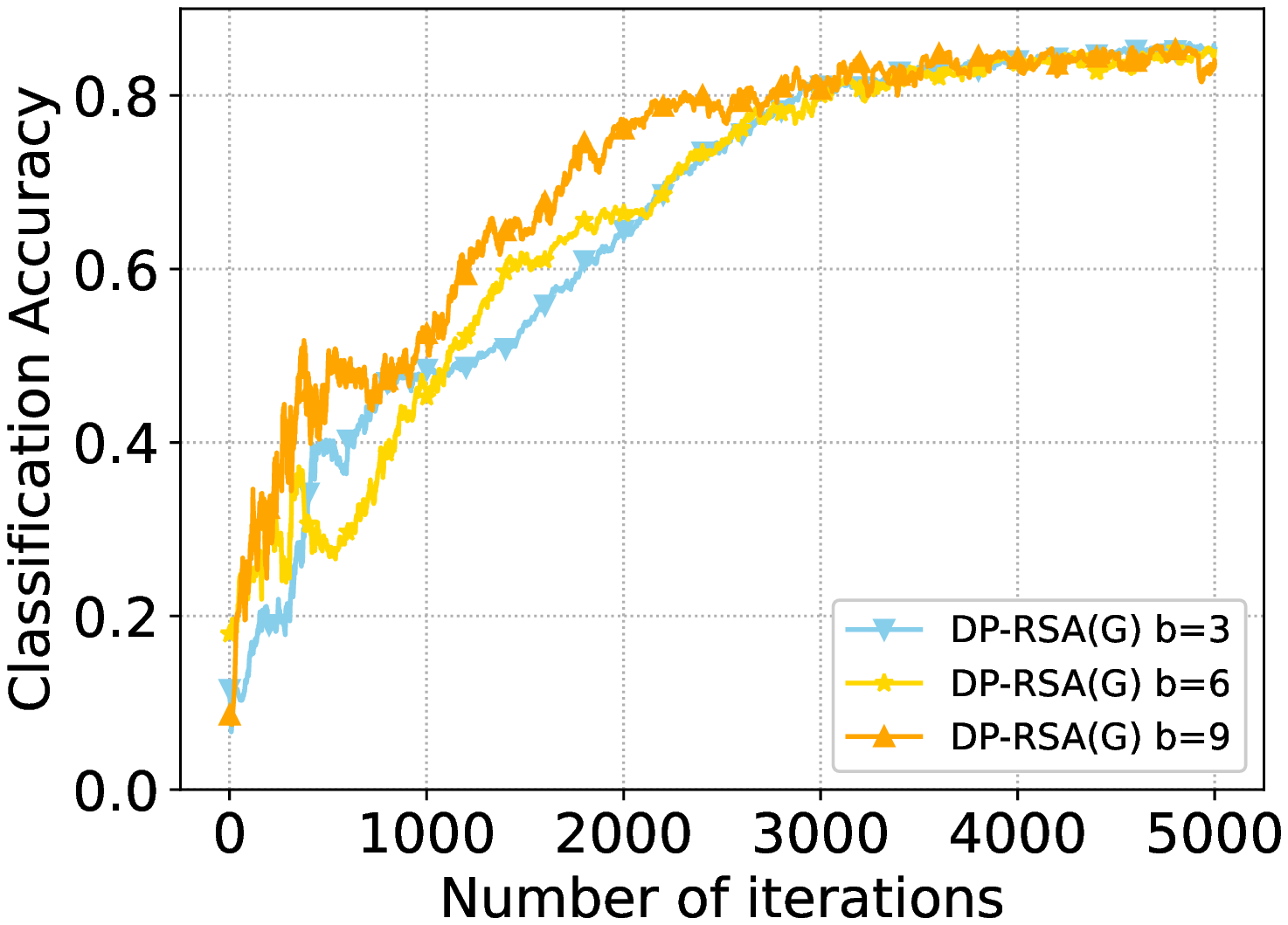}
    }
    \subfigure[CIFAR10 Dataset]{
    \includegraphics[width=0.30\textwidth]{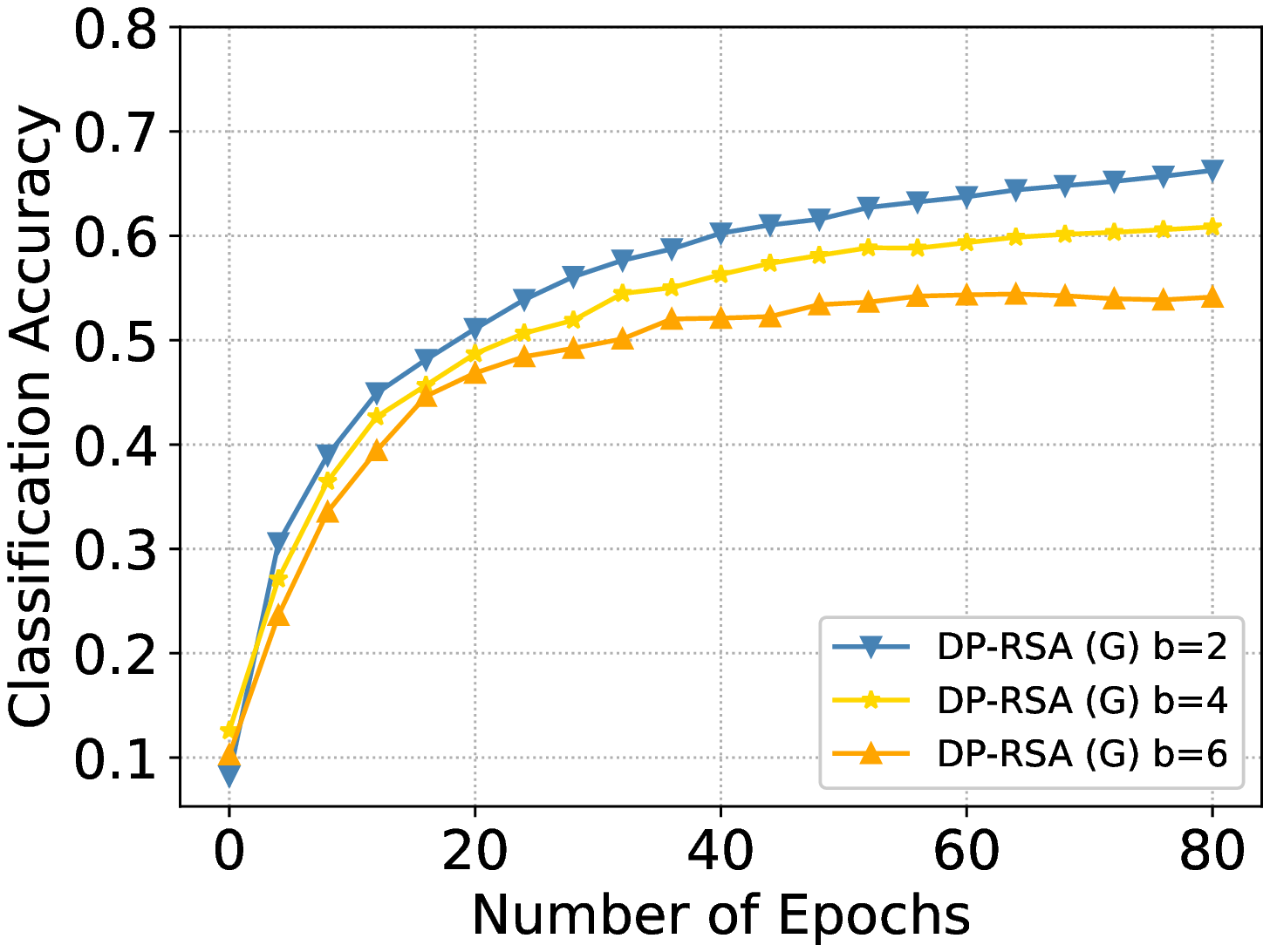}
    }
    \caption{Impact of number of Byzantine workers $b$, for Sample-Duplicating Attacks in the non-i.i.d. case.}
    \label{ByzantineSize}
\end{figure}

Here we investigate the impact of number of Byzantine workers $b$ on the learning performance. We use DP-RSA with the Sign-Gaussian mechanism and $\epsilon$ is 0.4. We consider Sample-Duplicating Attacks in the non-i.i.d. case. Fig. \ref{ByzantineSize} depicts the impact of number of Byzantine workers over the MNIST and CIFAR10 datasets, respectively. For the MNIST dataset we set the number of Byzantine workers to 3, 6 and 9, and for the CIFAR10 dataset we set the number to 2, 4, and 6. Other experimental settings are the same as those in the main text. For the smaller model over the MNIST dataset, the influence of Byzantine workers is not obvious even in the non-i.i.d. setting. However, for the large model trained over the CIFAR10 dataset, we can see more Byzantine workers will lead to much worse learning performance.

\begin{figure}[htb]
    \centering
    \subfigure[MNIST Dataset]{
    \includegraphics[width=0.30\textwidth]{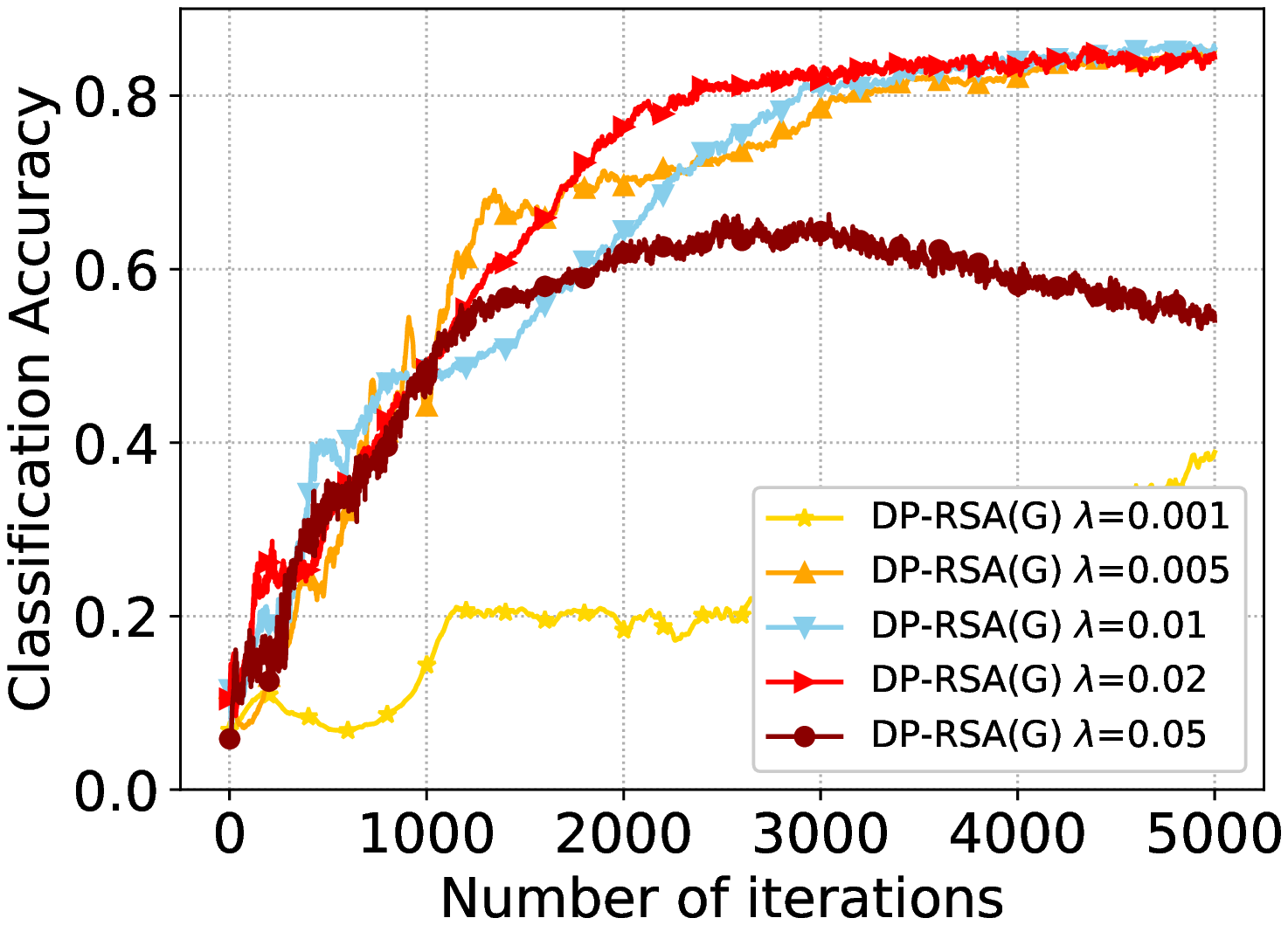}
    }
    \subfigure[CIFAR10 Dataset]{
    \includegraphics[width=0.30\textwidth]{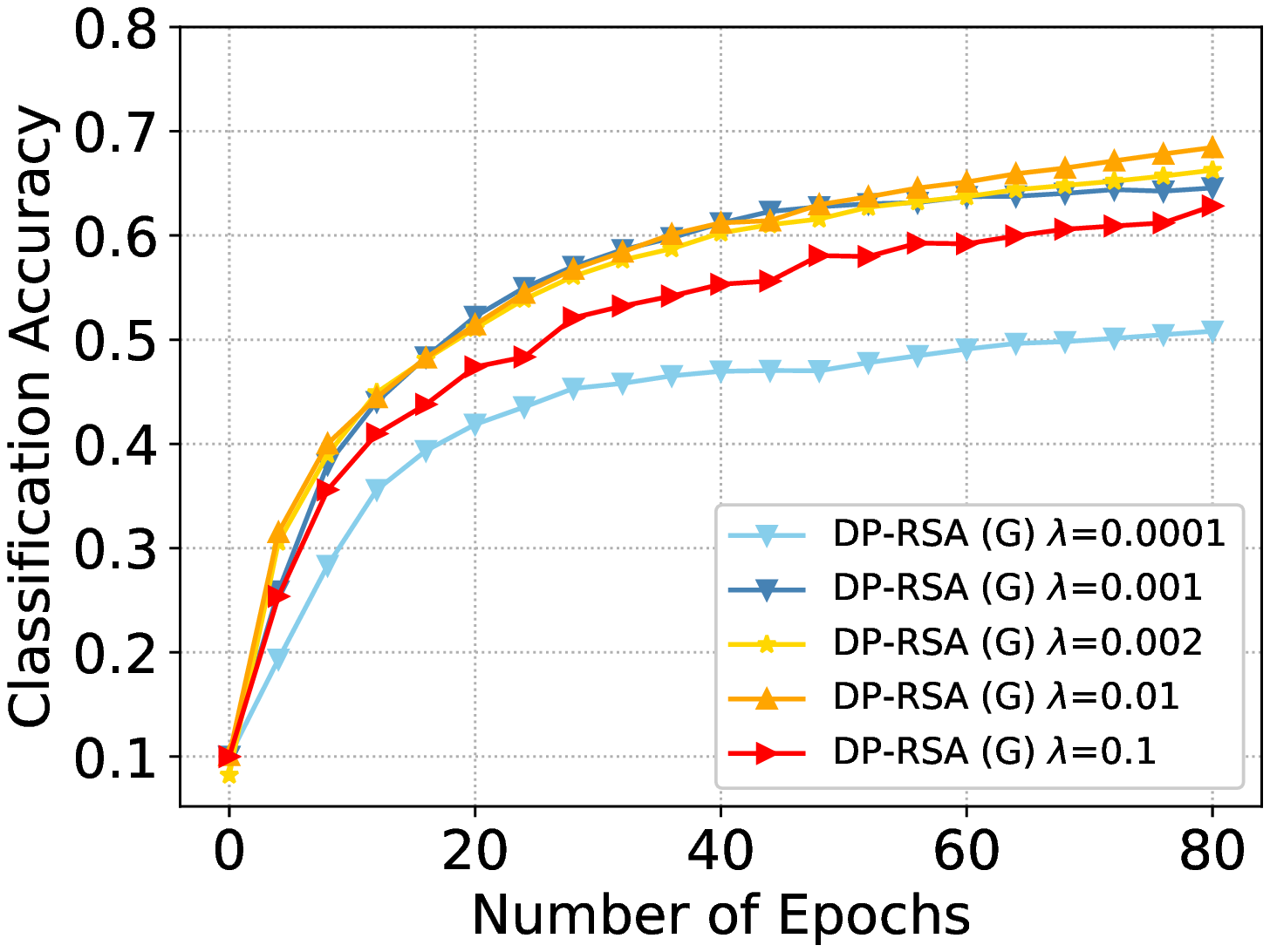}
    }
    \caption{Impact of penalty parameter $\lambda$, for Sample-Duplicating Attacks in the non-i.i.d. case.}
    \label{result_lambda}
\end{figure}

\subsection{Impact of Penalty Parameter $\lambda$}
Here we investigate the impact of penalty parameter $\lambda$ on the learning performance. We use DP-RSA with the Sign-Gaussian mechanism and $\epsilon$ is 0.4. We consider Sample-Duplicating Attacks in the non-i.i.d. case. Other experimental settings are the same as those in the main text.

Fig. \ref{result_lambda} shows the performance with different values of $\lambda$ on the MNIST and CIFAR10 datasets, respectively. For MNIST, we can see when $\lambda$ is 0.005, 0.01 and 0.02, the learning performance is satisfactory. However, when $\lambda$ is as small as 0.001 or as large as 0.05, the learning performance degrades sharply. When $\lambda$ is too small, the consensus constraints are unlikely satisfied, yielding remarkable learning error in the non-i.i.d. case. On the other hand, as we have seen in the theoretical analysis, the learning error is proportional to $\lambda^2$, and thus is unacceptable when $\lambda$ is too large. Therefore we suggest to choose a proper $\lambda$ to balance between the consensus and the learning error. For the more complicated model trained over CIFAR10, the influence of $\lambda$ is less obvious. In a wider range of $\lambda$, DP-RSA performs well. But for the cases that $\lambda$ is as large as $\lambda=0.1$ and as small as $\lambda=0.0001$, the performance just slightly degrades.

\end{document}